\documentclass{article} 
\usepackage{collas2024_conference,times}
\usepackage{easyReview}

\usepackage{amsmath,amsfonts,bm}

\def\eqref#1{equation~\ref{#1}}

\def\1{\bm{1}}

\DeclareMathAlphabet{\mathsfit}{\encodingdefault}{\sfdefault}{m}{sl}
\SetMathAlphabet{\mathsfit}{bold}{\encodingdefault}{\sfdefault}{bx}{n}

\usepackage{kantlipsum, lipsum}
\usepackage{dsfont}
\usepackage{hyperref}       
\usepackage{url}            
\usepackage{booktabs}       
\usepackage{amsfonts}       
\usepackage{nicefrac}       
\usepackage{microtype}      

\usepackage{amsmath}
\usepackage{amssymb}

\usepackage{times}
\usepackage{amsfonts} 
\usepackage{graphicx}
\usepackage{tabularx}
\usepackage{comment}
\usepackage{wrapfig}
\usepackage{amsmath}
\usepackage{amssymb}
\usepackage{array}
\usepackage{wrapfig,lipsum,booktabs}
\usepackage{hyperref}
\usepackage{booktabs}
\usepackage{multirow}
\usepackage{subfig}
\usepackage{setspace}
\usepackage{xcolor}

\usepackage{algorithm}
\usepackage{algorithmic}

\usepackage{hyperref}

\collasfinalcopy

\title{Diffusion Augmented Agents: A Framework for Efficient Exploration and Transfer Learning}

\author{Norman Di Palo\\
Imperial College London\\
London, UK\\
\texttt{normandipalo@gmail.com} \\
\And 
Leonard Hasenclever \thanks{Senior Author} \\
Google DeepMind \\
London, UK 
\And 
Jan Humplik \footnotemark[1]\\
Google DeepMind\\
London, UK 
\And 
Arunkumar Byravan \footnotemark[1] \\
Google DeepMind\\
London, UK 
}

\begin{document}



\graphicspath{{figures/}}


\maketitle

\begin{abstract}
We introduce Diffusion Augmented Agents (DAAG), a novel framework that leverages large language models, vision language models, and diffusion models to improve sample efficiency and transfer learning in reinforcement learning for embodied agents. DAAG hindsight relabels the agent's past experience by using diffusion models to transform videos in a temporally and geometrically consistent way to align with target instructions with a technique we call Hindsight Experience Augmentation. A large language model orchestrates this autonomous process without requiring human supervision, making it well-suited for lifelong learning scenarios. The framework reduces the amount of reward-labeled data needed to 1) finetune a vision language model that acts as a reward detector, and 2) train RL agents on new tasks. We demonstrate the sample efficiency gains of DAAG in simulated robotics environments involving manipulation and navigation. Our results show that DAAG improves learning of reward detectors, transferring past experience, and acquiring new tasks - key abilities for developing efficient lifelong learning agents. Supplementary material and visualizations are available on our website \href{https://sites.google.com/view/diffusion-augmented-agents/}{https://sites.google.com/view/diffusion-augmented-agents/}.

\end{abstract}

\section{Introduction}

\begin{wrapfigure}{r}{0.55\textwidth}
\vspace{-20pt}
    \begin{center}
    \includegraphics[width=0.55\textwidth]{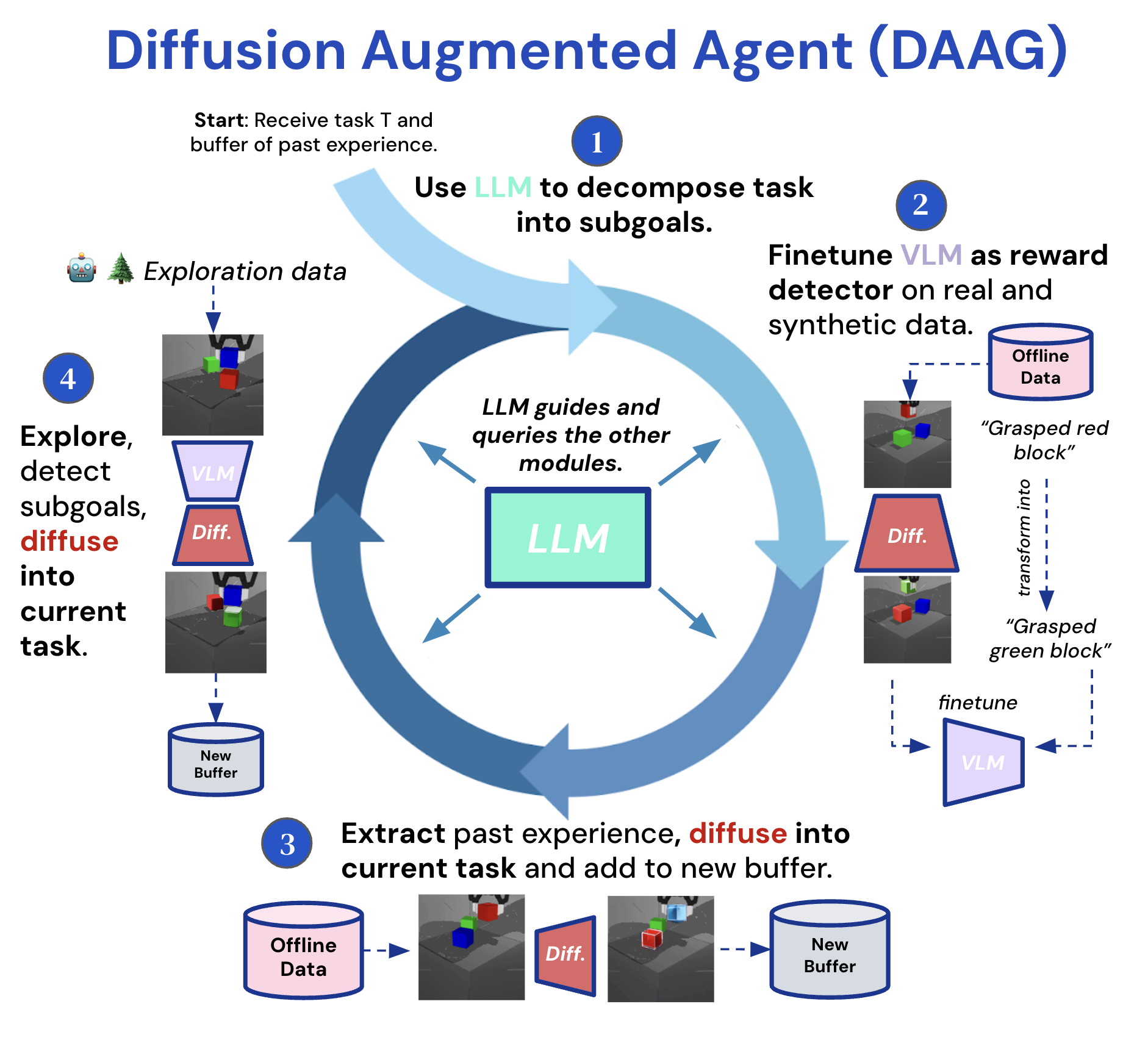}
    \end{center}
      \vspace{-10pt}

    \caption{An illustration of our proposed framework.}
    \label{fig:framework}
  \vspace{-10pt}
\end{wrapfigure}

The most recent notable breakthroughs in AI have come from the combination of large models trained on enormous datasets \citep{firoozi2023foundation, brown2020language, hoffmann2022training, reed2022generalist, geminiteam2023gemini}. However, despite efforts to scale up data collection \citep{embodimentcollaboration2023open, reed2022generalist, bousmalis2023robocat}, data in embodied AI settings is still prohibitively scarce because such agents need to interact with physical environments where sensors and actuators present major bottlenecks \citep{cabi2020scaling, lee2022beyond}.

This data scarcity issue is especially pronounced in reinforcement learning scenarios, where rewards are often sparse or completely absent in realistic settings \citep{ecoffet2021goexplore}. Overcoming these challenges requires developing agents that can learn and adapt efficiently from limited experience. We hypothesize that embodied agents can achieve greater data efficiency by leveraging past experience to explore effectively and transfer knowledge across tasks (e.g. \citep{andrychowicz2017hindsight}). In particular, we are interested in enabling agents to autonomously set and score subgoals, even in the absence of external rewards, and to repurpose their experience from previous tasks to accelerate learning of new tasks.

In this paper, we address these questions using foundation models pretrained on internet-scale datasets \citep{firoozi2023foundation}. Through an interplay of vision, language, and diffusion models \citep{radford2021clip, alayrac2022flamingo, brown2020language, stablediffusion}, we enable the agent to more effectively reason about its tasks, interpret its environment and past experience, and manipulate the data it collects to repurpose it for novel tasks and objectives. Importantly, DAAG operates autonomously without the need for human supervision, making it particularly well-suited for lifelong reinforcement learning scenarios.

In Figure \ref{fig:framework} we illustrate our framework from a high-level perspective. A large language model (LLM) acts as the main controller, or \textit{brain}, querying and guiding a vision language model (VLM) and a diffusion model (DM) and the high-level behavior of the agent. We call our framework \textbf{D}iffusion \textbf{A}ugmented \textbf{Ag}ent (\textbf{DAAG}).

\begin{figure*}[t!]
    \centering
    \includegraphics[width=.99\textwidth]{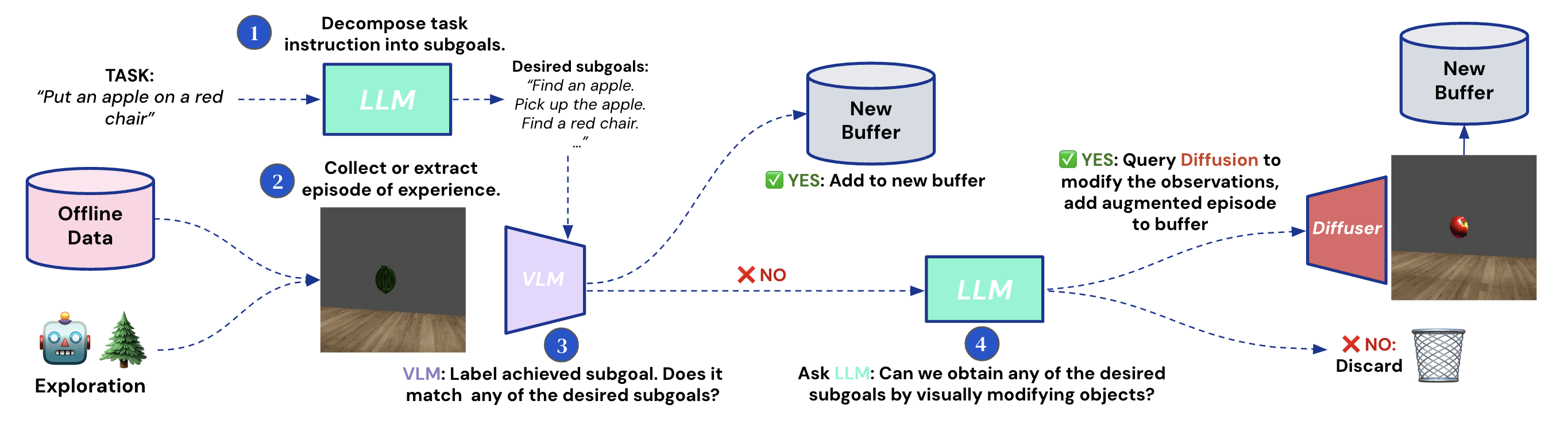}
    \centering
    \caption{An example of the Hindsight Experience Augmentation pipeline, with which new or past experience can be optionally modified and added to a new buffer.}
    \label{fig:hea}
\end{figure*}

Through a series of experiments on different environments, including robot manipulation and navigation, we empirically demonstrate that our proposed framework improves the agent's performance on a series of key abilities: \textbf{1)} \textit{autonomously computing rewards} for both seen and unseen tasks by fine-tuning a vision language model on data augmented with synthetic samples generated by a diffusion model; \textbf{2)} \textit{more efficiently exploring and learning new tasks} by designing and recognising useful sub-goals for the given task, and repurposing otherwise failed trajectories by modifying the recorded observations via diffusion models; \textbf{3)} \textit{effectively transferring previously collected data to new tasks} by both extracting related data and repurposing other trajectories through the use of diffusion models. In Figure \ref{fig:hea}, we illustrate how our method can repurpose agent's experience via diffusion augmentation. We propose a diffusion pipeline that improves geometrical and temporal consistency to modify part of videos collected by agents (Fig. \ref{fig:diffusion_pipeline}), a novelty in the field of reinforcement learning to the best of our knowledge.


\section{Related Work}

\textbf{Foundation Models in Embodied AI} The rapid evolution of foundation models \citep{bommasani2022opportunities} such as large language models (LLMs) \citep{brown2020language, hoffmann2022training, geminiteam2023gemini} and vision language models (VLMs) \citep{openai2023gpt4, geminiteam2023gemini, alayrac2022flamingo, radford2021clip} has sparked significant interest from the robotics and embodied AI research community in recent years. As these models have demonstrated increasingly sophisticated abilities in language understanding, reasoning, commonsense knowledge, and visual perception, researchers have leveraged their potential as building blocks for intelligent agents.

Many methodologies for integrating these models into robotics systems were proposed in recent months. \citep{wang2023survey, wang2024large, firoozi2023foundation}. \citep{ahn2022saycan, liang2023code, dipalo2023unified, huang2022language} proposed the use of LLMs as high-level planner, able to decompose an instruction into a series of short horizon sub-goals. These models than employ different modalities to ground such textual plans into actions. \citep{huang2023voxposer, yu2023language} use LLMs and VLMs to instead obtain a reward function, given a textual instruction, that can be then used by an external optimiser to compute a trajectory. \citep{brohan2023rt2, kwon2023language} use vision and language models to directly output executable actions given a textual description of a task. \citep{xiao2023robotic, dipalo2023unified} use VLMs as reward detectors/task classifiers. In this work, we build upon the framework proposed in \citep{dipalo2023unified}, using an LLM to decompose long horizon plans into a sequence of subgoals. However, we extend it by giving the LLM the ability to query a diffusion model to modify and augment visual observations autonomously, unlocking faster learning and more effective transfer.

\textbf{Image Generation with Diffusion Models} The rapid evolution of generative AI, beyond language-based application, also saw the exponential growth of image generation models, largely based on diffusion processes \citep{dhariwal2021diffusion, stablediffusion, ho2020denoising, ramesh2021zeroshot}. Mostly, these models are conditioned via text, specifying the desired output image in natural language. Beyond generating images from scratch, these models have also been employed to modify images via in-painting, i.e. the completion of a specific, masked part of an original image. To provide more fine-grained control of the generation process, recent works have enabled diffusion models with the ability to be conditioned on modalities other than text:  \citep{zhang2023adding} demonstrated the use of depth maps, canny edges, segmentation masks and more in addition to textual instructions to guide and constrain the geometrical and visual aspect of the final outputs. We largely based our proposed diffusion pipeline on these results, as ensuring the augmented observations respect the original geometries of objects and environment is fundamental in embodied applications.
Image-based diffusion models have also been recently extended into the temporal domain, outputting short videos \citep{blattmann2023stable} instead of single frames. Training such models is however particularly challenging, and they do not yet benefit from the fine grained control abilities mentioned above. Notably, \citep{khachatryan2023text2videozero} demonstrated that image diffusion models can be repurposed to generate videos simply by constraining the original noise maps used to start the backward diffusion process.

\textbf{The Use of Diffusion Models in Robotics}  While notable research has been conducted on the use of diffusion models as a way to represent policies \citep{chi2023diffusion} or generate additional experience for the agent as low-level states \citep{lu2023synthetic}, in this work we will focus on their use in the visual domain as text-conditioned image generators. As visual perception is fundamental for robotics and embodied AI, the recent literature investigated methodologies to integrate the image generation abilities of such models to empower agents \citep{zhu2024diffusion}. To augment the visual experience collected by a robot, \citep{chen2023genaug, mandi2023cacti, yu2023scaling} propose the use of diffusion models to either modify the background to increase robustness to distractors, or repurpose trajectory for new tasks by modifying the manipulated objects into new ones of interest \citep{yu2023scaling}. While the latter is similar to our proposed approach, there are some fundamental differences: \citep{yu2023scaling} is based on an imitation learning pipeline, where a human operator labels the task demonstrated and the new task to generate via diffusion. Our method, designed and revolving around the abilities of LLMs, is entirely autonomous both in terms of detecting the accomplished tasks (via the VLM) and proposing and generating augmentations (via the LLM querying the diffusion model), therefore being more suited for (lifelong) reinforcement learning scenarios, where human supervision is not always present. Additionally, we propose a diffusion pipeline that is both geometrically and temporally consistent when modifying video, differently from \citep{chen2023genaug, mandi2023cacti, yu2023scaling}. Ensuring temporal and geometrical consistency is beneficial when dealing with embodied environments, as we later demonstrate.

\begin{figure*}[t!]
    \centering
    \includegraphics[width=.7\textwidth]{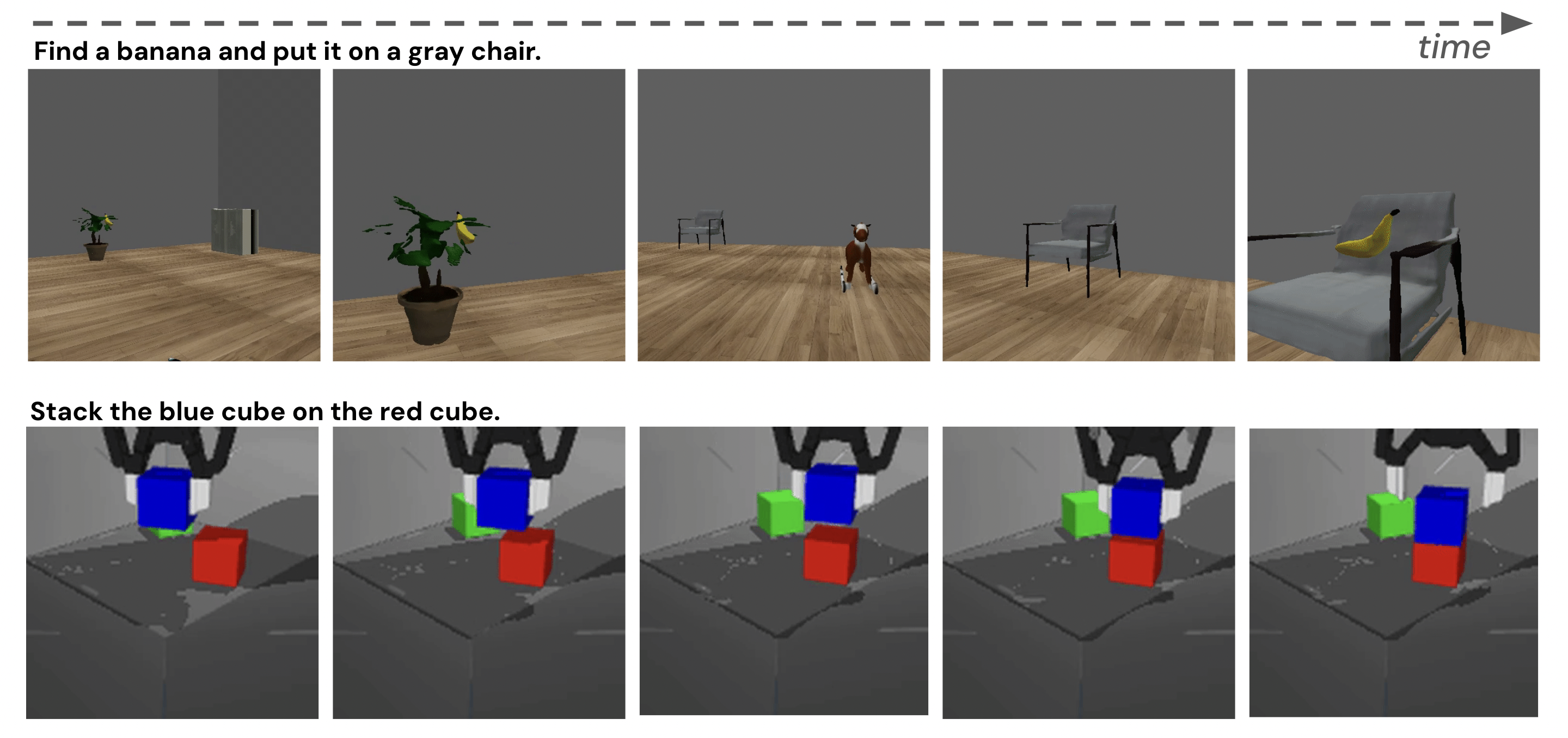}
    \centering
    \caption{Two example instances of tasks we use in our investigation.}
    \label{fig:two_tasks}
\end{figure*}

\textbf{Experience Re-Use and Transfer} The need for vast amounts of experience data to learn policies \citep{reed2022generalist, bousmalis2023robocat} inspired the research community to propose methods and strategies to re-use experience collected for different, previous tasks. Hindsight Experience Replay \citep{andrychowicz2017hindsight} proposes a method to re-use episodes that solve tasks different from the desired task. If asked to solve task $\mathcal{T}_n$, but solving instead task $\mathcal{T}_m$ during exploration, \citep{andrychowicz2017hindsight} proposed to relabel the episode as if the desired task was $\mathcal{T}_m$, therefore re-purposing a trajectory as a success for a different task. Differently, when in the same situation, our method modifies the collected observations that solve task $\mathcal{T}_m$ to synthetically generate an episode that would have solved the desired task $\mathcal{T}_n$. This means we can generate data for a task \textit{without the need to effectively solve it during exploration}.
\citep{bousmalis2023robocat} repurposes data from various task by learning a single, goal-conditioned policy, therefore extracting learning signal from any episode solving any task. However, in our experiments we demonstrate that trying to learn a single, goal-conditioned policy is not as effective as explicitly generating and training on data for the specific task we are interested in.
\citep{tirumala2023replay} demonstrated the effectiveness of re-using collected experience from past runs and experiments to bootstrap learning: in this work, we first extract only relevant data, as in \citep{dipalo2023unified}, through the use of vision and language models, and additionally generate bespoke new data from past data for the new task at hand, demonstrating improved learning efficiency.

\section{Method}
\label{sec:method}

\begin{figure*}[t!]
    \centering
    \includegraphics[width=.99\textwidth]{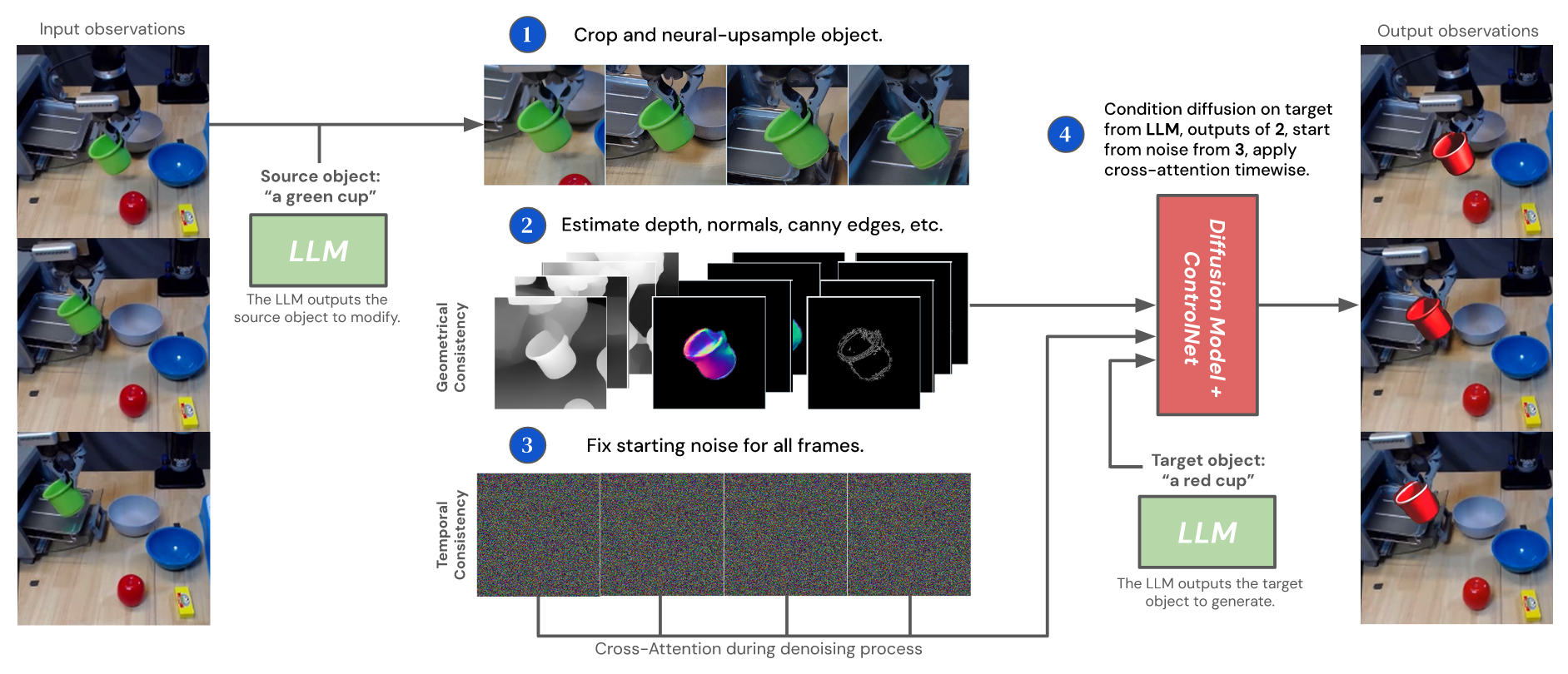}
    \centering
    \caption{An illustration of our diffusion pipeline, highlighting the geometrical and temporal consistency obtained by combining the methodologies in \citep{zhang2023adding} and \citep{khachatryan2023text2videozero}.}
    \label{fig:diffusion_pipeline}
\end{figure*}

\subsection{Preliminaries}
 We formalise our environments as Markov Decision Processes (MDPs): the environment and the agent, at each timestep $t$, are in a state $s \in \mathcal{S}$. From that state, the agent receives a visual observation $o \in \mathcal{O}$, and can execute an action $a \in \mathcal{A}$. During each episode, the agent receives an instruction, which is a description of the task to execute in natural language $\mathcal{T}$. The agent can receive a reward $r = \text{+} 1$ at the end of the episode if the task is successfully executed.

In this work, beyond learning new tasks in isolation, we study our framework's ability to learn tasks in succession in a lifelong fashion. Therefore, the agent stores interaction experiences in two buffers: the current task buffer that we call \textit{new buffer} $\mathcal{B}_{n}$: this buffer is initialised at the beginning of each new task. There is then an \textit{offline lifelong buffer} $\mathcal{B}_{ll}$: the agent stores all episodes from all tasks in this buffer, regardless of their success. The latter is therefore an ever-growing buffer of experiences the agent can then use to bootstrap learning of new tasks.

\textbf{Large Language Model:} We use a large language model to orchestrate the behaviour of the agent and the use of the vision language model (VLM) and diffusion model. The LLM receives a textual instruction and data and outputs textual responses. In our work, we leverage the LLM's ability to decompose tasks into subgoals, compare the similarity of different tasks/instructions, and query the VLM and diffusion model. We parse the output of the LLM to obtain the exact string we need for each use case. In the Supplementary Material on our website, we show the way we design the prompt to guide the textual generation of the LLM and simplify the final parsing of its output. 

\textbf{Vision Language Model: }The VLM we use is \textbf{CLIP} \citep{radford2021clip}, a contrastive model. CLIP is composed of two branches: an image branch $\phi_{\text{image}}$ and a textual branch $\phi_{\text{text}}$. They respectively take as input visual observations and textual descriptions, outputting embedding vectors of the same size $y_{\text{t,im}} = \phi_{\text{image}}(o_t)$,  $y_{\text{g,txt}} = \phi_{\text{text}}(\mathcal{T}_g)$. The peculiarity of the output embeddings is the following: their cosine similarity $s_{g,t} = cs(y_{\text{t,im}}, y_{\text{g,txt}})$ implicitly represents how well the text $\mathcal{T}_t$ describes the observation $o_t$. Following \citep{dipalo2023unified}, we consider that the VLM labels an observation $o_t$ as being a goal observation for task $\mathcal{T}_g$ if $s_{g,t} > \delta$, where $\delta$ is a threshold computed during training. More details are presented in the Supplementary Material on our website. Therefore, given a set of goal tasks  $\mathcal{T}_{0:G}$, for each new observation CLIP computes a new  $s_{g,t} = cs(y_{\text{t,im}}, y_{\text{g,txt}})$ and if the threshold is surpassed, labels the observation as achieving the task at hand.

As CLIP needs to be explicitly given as an input a textual description, in addition to the image, we need to manually provide a set of tasks we are interested in detecting while the agent is exploring. Briefly, we adopt two strategies: first, we keep a list of all the tasks that were given to the agent up to that point, and all the relative subgoals obtained by the LLM. Additionally, we can autonomously propose possible subgoals by giving to the LLM the aforementioned list and a list of objects present in the environment. These techniques allow us to have a list of tasks that the agent may randomly achieve at test time.

\textbf{Diffusion Pipeline:} A core aspect of our work is modifying visual observations through language-instructed diffusion models. The goal of our diffusion pipeline is to take an observation $o_t$ or a temporal series of observations recorded by the agent $o_{t:t+H}$ and visually modify one or more objects present in the observation(s), while enforcing geometrical and temporal consistency. 

\begin{wrapfigure}{r}{0.4\textwidth}

    \begin{center}
    \includegraphics[width=0.4\textwidth]{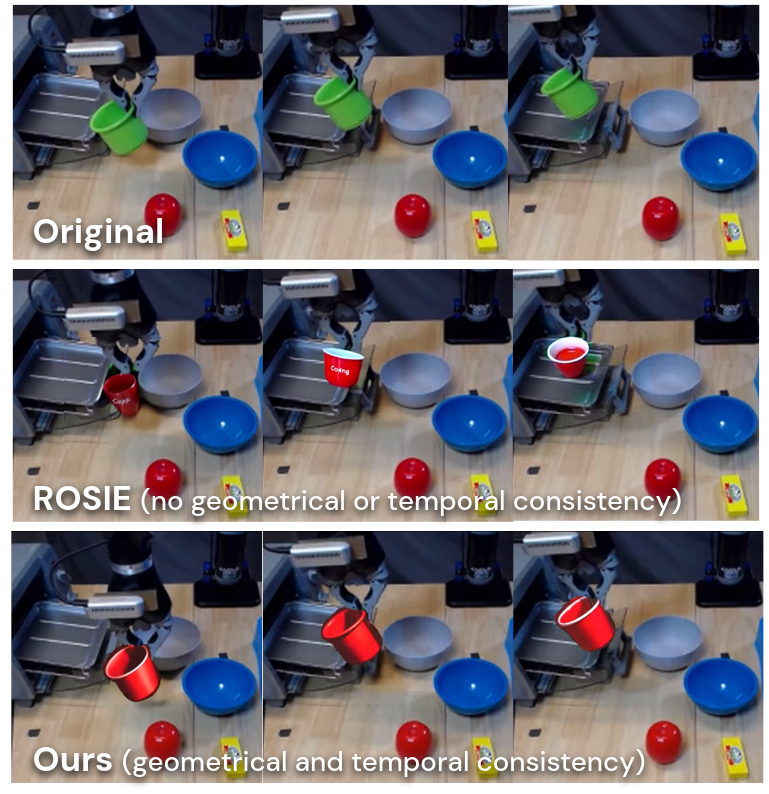}
    \end{center}
    \vspace{-10pt}

    \caption{Outputs of the method proposed in ROSIE \citep{yu2023scaling} and our diffusion pipeline when asked to "\textit{swap a green cup with a red cup}".}
    \label{fig:rosie_vs_ours}
    \vspace{-10pt}
\end{wrapfigure}

To  tackle the former, the diffusion pipeline receives as input the observation(s) to be modified, a textual description of the object to modify, or \textit{source object} $\text{obj}_s$, and a  description of the object to generate in place of the source, or \textit{target object} $\text{obj}_t$. The former is used to localise, crop and mask the source object to perform in-painting, while the latter is used by the diffusion model itself to guide the image generation. The pipeline receives, additionally, a series of visual inputs to condition the image generation, all computed from the original observation(s): depth maps $o_\text{depth}$, canny edges $o_\text{canny}$, and normals maps $o_\text{normals}$. We compute these from RGB observations via a series of off-the-shelf models we list in the Supplementary Material. The latter are used, through the ControlNet architectures  \citep{zhang2023adding}, to ensure that the generated objects respect the geometry of the original objects. While the use of each is optional, their combined use improves the overall results. The overall process can be described as $ \hat{o}_{t:t+H} = \text{Diff}(o_{t:t+H}, \text{obj}_s, \text{obj}_t, o_\text{depth}, o_\text{canny}, o_\text{normals})$, where $\hat{o}_t$ represents an observation modified via diffusion from the original observation $o_t$. Fig. \ref{fig:diffusion_pipeline} represent an illustration of the entire pipeline.

To improve temporal consistency, we apply the technique proposed in \citep{khachatryan2023text2videozero}: when applying diffusion to $N$ frames, we 1) fix the initial noise map for all $N$ instead of sampling different ones (the results will still be different as the ControlNet inputs will be different), and 2) add \textbf{temporal cross-attention} to the diffusion process: all the frames can therefore attend to all other frames during the backward diffusion process for image generation. This does not require any architectural change, nor retraining the model. In Figure \ref{fig:rosie_vs_ours}, we provide a visual comparison of the outputs of the method proposed in \citep{yu2023scaling} and our proposed pipeline. The former does not keep object poses and aspect consistent over frames, therefore invalidating the hypothesis that applying the same actions $a_{0:T}$ to the modified observations $\hat{o}_{0:T}$ would have led to successful completion of the new task.


Now that we introduced the main components of DAAG, we will describe the way they interoperate in our framework. 

\subsection{Finetune, Extract, Explore: The Diffusion Augmented Agent Framework}

\textbf{Finetuning VLMs as Reward Detectors on Diffusion Augmented Data}: \label{sec:method_finetune} VLMs can be effectively employed as reward detectors, conditioned on a language-defined goal and a visual observation. However, as demonstrated by recent works \citep{dipalo2023unified, xiao2023robotic}, to be accurate they often need to be finetuned on labelled data gathered in the target environment, for the desired tasks. This is a time-consuming task that furthermore requires human effort for each new task to be learned, hindering the ability of the agent to autonomously learn many tasks in succession in a lifelong fashion. With our framework we tackle this challenge by finetuning the VLM on previously collected observations. Given a dataset $\mathcal{D}$ of observations $o_i$, each paired with a label $\mathcal{T}_i$, and a new goal task expressed in natural language, $\mathcal{T}_g$, we extract all observations whose caption $\mathcal{T}_i$ is similar enough to $\mathcal{T}_g$, or such that a visual modification of the corresponding observation $o_i$ would transform it into a fitting observation $\hat{o_i}$ for $\mathcal{T}_g$: for example, given the goal description "\textit{The robot is grasping the red cube}", an observation with caption "\textit{The robot is grasping the blue cube}" can be modified by visually swapping \textit{red cube} with \textit{blue cube} through a controlled diffusion process. In DAAG, the LLM autonomously selects the fitting observations $o_i$ from $D$ by comparing their caption $\mathcal{T}_i$ with the goal caption $\mathcal{T}_g$: if a swap is considered possible, the LLM instructs the DM with the source object to modify and the target object to add (in the previous example $\{ \textit{blue cube, red cube}\}$ respectively). This process is illustrated in Figure \ref{fig:hea}. Through this process, we finetune the VLM to act as a success detector for all the subgoals $\mathcal{T}_{0:G}$ in which the task at hand was decomposed by the LLM.

\textbf{Efficient Learning and Transfer via Hindsight Experience Augmentation}: After each episode collected on any task it encounters, an agent collects a series of observations and actions $E_n = \{o_t, a_t \}_{t=0}^T$. We store every episode, regardless of the final outcome, in the lifelong buffer $\mathcal{B}_{ll}$ \citep{cabi2020scaling}. When learning a new task, the agent receives a task instruction in textual form $\mathcal{T}_g$ and decomposes it into a series of subgoals  $\mathcal{T}_{0:G}$ via the LLM. 
Normally, the agent can extract a learning signal only from episodes that are collected via exploration or from past experience stored in $\mathcal{B}_{ll}$ if there are rewards associated to it; these rewards can be either from the VLM or be an external reward from the environment. In DAAG, we aim to maximise the number of episodes from which the agent can learn to tackle a new task, even if it does not achieve any of the desired subgoals. We do this through a process we call \textbf{Hindsight Experience Augmentation (HEA)}.

\begin{figure*}[t!]
    \centering
    \includegraphics[width=.99\textwidth]{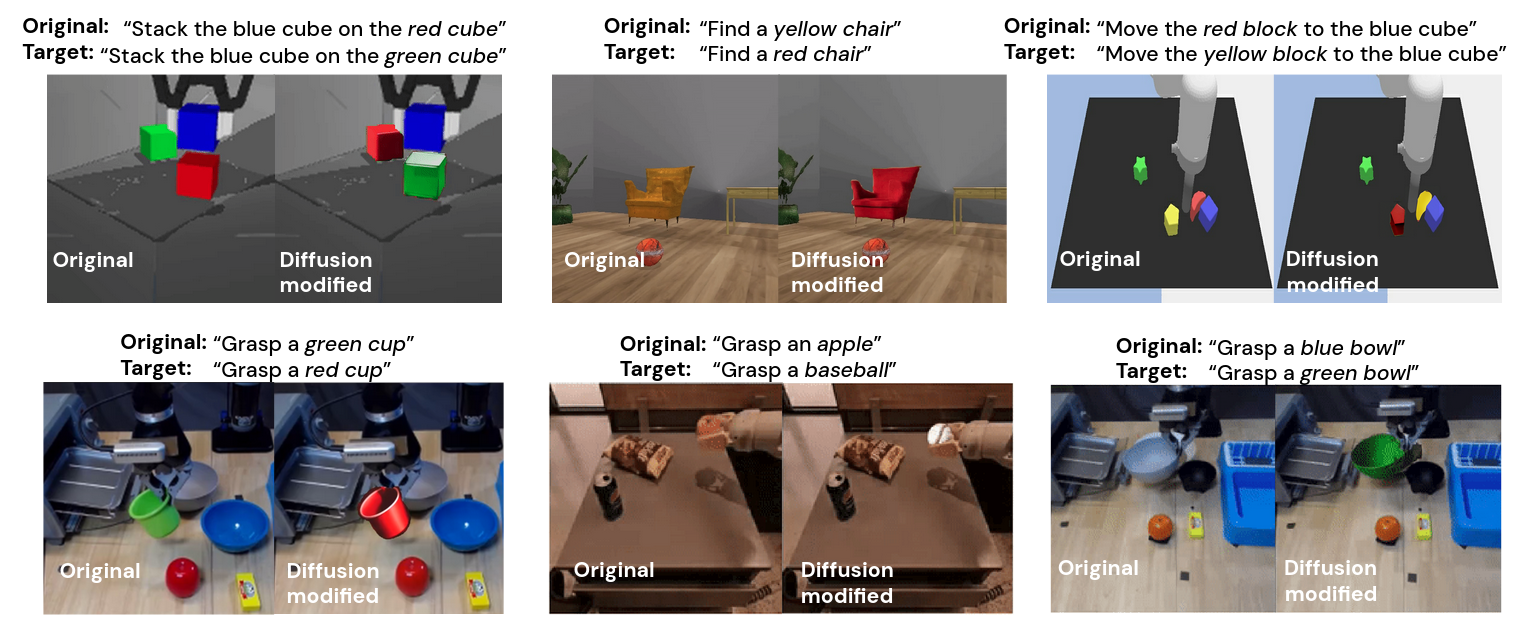}
    \centering
    \caption{Examples of original observations and diffusion modified observations, showing the original achieved goal and the desired goal, taken from our environments and from the Open X-Embodiement dataset \citep{embodimentcollaboration2023open}.}
    \label{fig:diff_modified_examples}
    \vspace{-15pt}
\end{figure*}

Given an episode of experience $E_i = \{o_t, a_t \}_{t=0}^T$, we use the VLM to label what possible subgoals have been achieved by the agent, as described in the Preliminaries (more details on this phase are presented in the Suppl=['9o0]
-\=ementary Material). If any matches a desired subgoal, we add this episode to the new, current task buffer $\mathcal{B}_n$. This process emulates the framework proposed in \citep{dipalo2023unified}. However, if no match is present, instead of discarding the episode, we query the LLM to ask if the achieved subgoal(s) can match any of the desired subgoals by swapping/visually modifying any of the objects, e.g. matching \textit{“The red cube is stacked on the blue cube”} with  \textit{“The green cube is stacked on the blue cube”} by swapping \textit{red cube} with \textit{green cube}. When a swap is identified as possible, the LLM queries the diffusion model to modify the observations of the episode up to the achieved sub-goal $[o_0, \dots, o_{T_g}]$ into $[\hat{o}_0, \dots, \hat{o}_{T_g}]$, and finally adds the modified observations and the original actions in the experience buffer $\mathcal{B}_n \leftarrow [(\hat{o}_0, a_0), \dots, (\hat{o}_{T_g}, a_{T_g})]$.

Through HEA, we can synthetically increase the number of successful episodes the agent can store in its buffers and learn from. This allows to effectively re-use as much data gathered by the agent as possible, substantially improving efficiency especially when learning multiple tasks in succession, as we will describe later. While previous methods have proposed the use of diffusion-based image generation or augmentation to synthesise additional data for learning policies \citep{yu2023scaling, mandi2023cacti, chen2023genaug}, our method is the first to propose an entire autonomous pipeline, independent from human supervision, and that leverages geometrical and temporal consistency to generate consistent augmented observations.

When learning a new task, the agent first applies HEA to all the episodes stored in its lifelong buffer $\mathcal{B}_{ll}$ to start with a non-empty new task buffer $\mathcal{B}_{n}$. It then trains a policy on this data to kickstart exploration, and subsequently applies HEA to any new episode of experience gathered via exploration. By dividing a goal into shorter-horizon subgoals via the LLM, and more efficiently learning to tackle those via HEA, our agent quickly learns to guide exploration, as it learns to solve the various subgoals that lead to the completion of the task. In a robotic stacking scenario, for example, DAAG efficiently learns to pick up the first object to stack. By consistently solving that first step during exploration, it is more likely to also randomly complete the task by placing it on top of the target object.

We will shed light on the effect of both of these phases on learning efficiency in the Experiments section. The entire pipeline is illustrated in Fig. \ref{fig:hea}.

\section{Experiments}
Our framework, \textbf{DAAG}, proposes an interplay between LLMs, VLMs and diffusion models to tackle three principal challenges in agents that learn in a lifelong fashion: \textbf{1)} finetuning a new reward/sub-goals detection model, \textbf{2)} extracting and transfering past experience for new tasks and \textbf{3)} efficiently exploring new tasks. To thoroughly investigate the benefits of DAAG on these three challenges, we designed and ran a series of experiments that individually measure the contribution and benefits of our method on each of these scenarios. The rest of the section is therefore divided into three main subsections, where each of these challenges is investigated and results are demonstrated and analysed: \textbf{1)} can DAAG finetune VLMs as reward detectors for novel tasks? \textbf{2)} can DAAG explore and learn new tasks more efficiently? \textbf{3)} can DAAG more effectively learn tasks in succession, transferring experience from past tasks?

\subsection{Experimental Setup}
We use three different environments to measure the performance of DAAG on the aforementioned challenges: \textbf{1)} a robot manipulation environment, \textbf{RGB Stacking} \citep{lee2022beyond}, where a robot arm is tasked with stacking colored cubes into a goal configuration. The action space $\mathcal{A}$ is composed of a target and goal pick and place position, represented as a pair of $\mathbb{R}^2$ numbers, and the observation space $\mathcal{O}$ is composed of RGB visual observations captured from a fixed shoulder camera. \textbf{2)} a navigation environment, \textbf{Room}, inspired by \citep{deepmindinteractiveagentsteam2022creating}, where an agent navigates in a room filled with objects and furniture, and is tasked with picking up and placing a goal object on a goal chair. The action space $\mathcal{A}$ is composed of a forward velocity and rotational velocity input represented as $\mathbb{R}^2$. We assume the agent can pick up an object automatically by moving in close proximity to it, and equally place it on a target object when sufficiently close to it. The observation space $\mathcal{O}$ is composed of RGB visual observations captured from a first-person view camera. \textbf{3)} a non-prehensile manipulation environment, \textbf{Language Table} \citep{lynch2022interactive}, where a robot can push colored blocks on a table to move them to a goal configuration. The action space $\mathcal{A}$ is composed of $x$ and $y$ end-effector velocity inputs as $\mathbb{R}^2$. The observation space $\mathcal{O}$ is composed of RGB visual observations captured from a fixed shoulder camera. Goals for all environments are provided as \textit{natural language instructions}. 

As a policy learning algorithm, we use Self-Imitation Behavior Cloning \citep{chen2021decision, dipalo2023unified, oh2018self} on all the episodes stored in the buffer $\mathcal{B}_n$, where the agent collects all successful episodes.

As a large language model, we use Gemini Pro \citep{geminiteam2023gemini}. As a VLM, we use CLIP \texttt{ViT-B/32} \citep{radford2021clip}. As a diffusion model, we use Stable Diffusion 1.5 \citep{stablediffusion}, with ControlNet \citep{zhang2023adding}.

Detailed hyperparameters and values of constant are listed in the Supplementary Material.

\subsection{Can DAAG Finetune VLMs as Reward Detectors for Novel Tasks?}
\label{sec:exp_finetune}

In this section we evaluate the ability of DAAG to obtain effective reward detectors for new tasks by finetuning VLM on past experiences collected by the agent being augmented through a diffusion pipeline, as explained in \ref{sec:method}.

\begin{figure*}[t!]
    \centering

    \includegraphics[width=.32\textwidth]{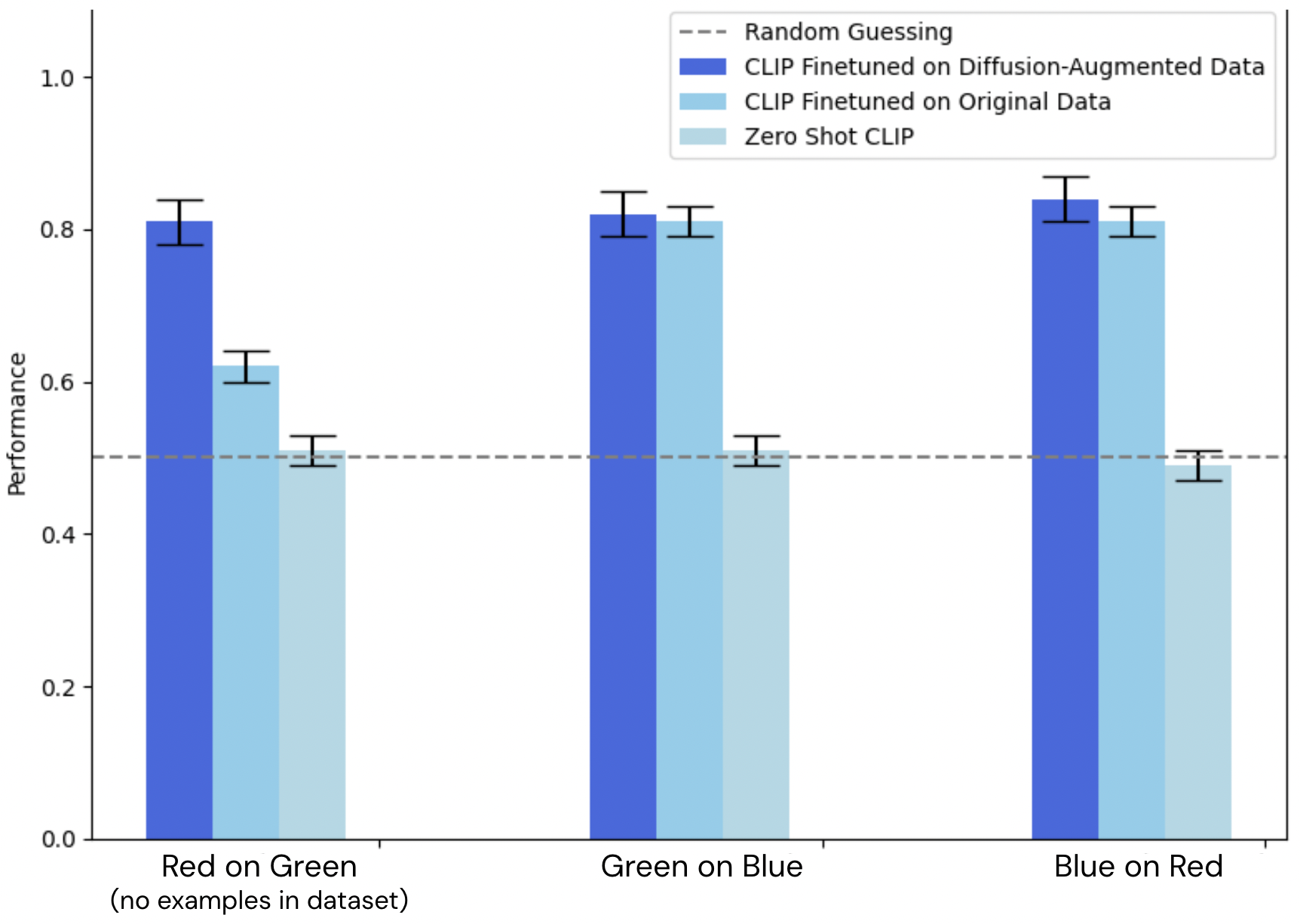}
        \includegraphics[width=.318\textwidth]{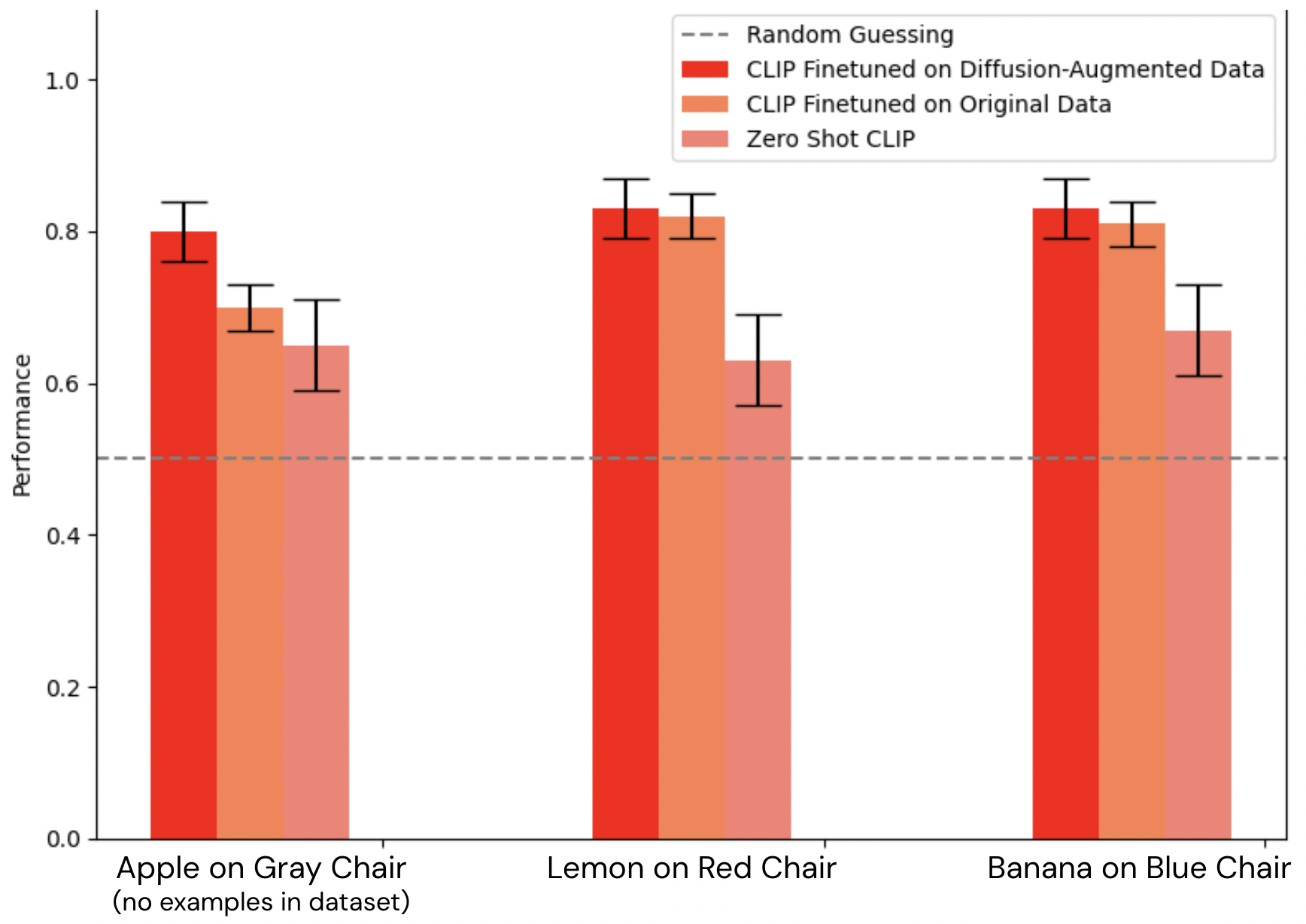}
            \includegraphics[width=.32\textwidth]{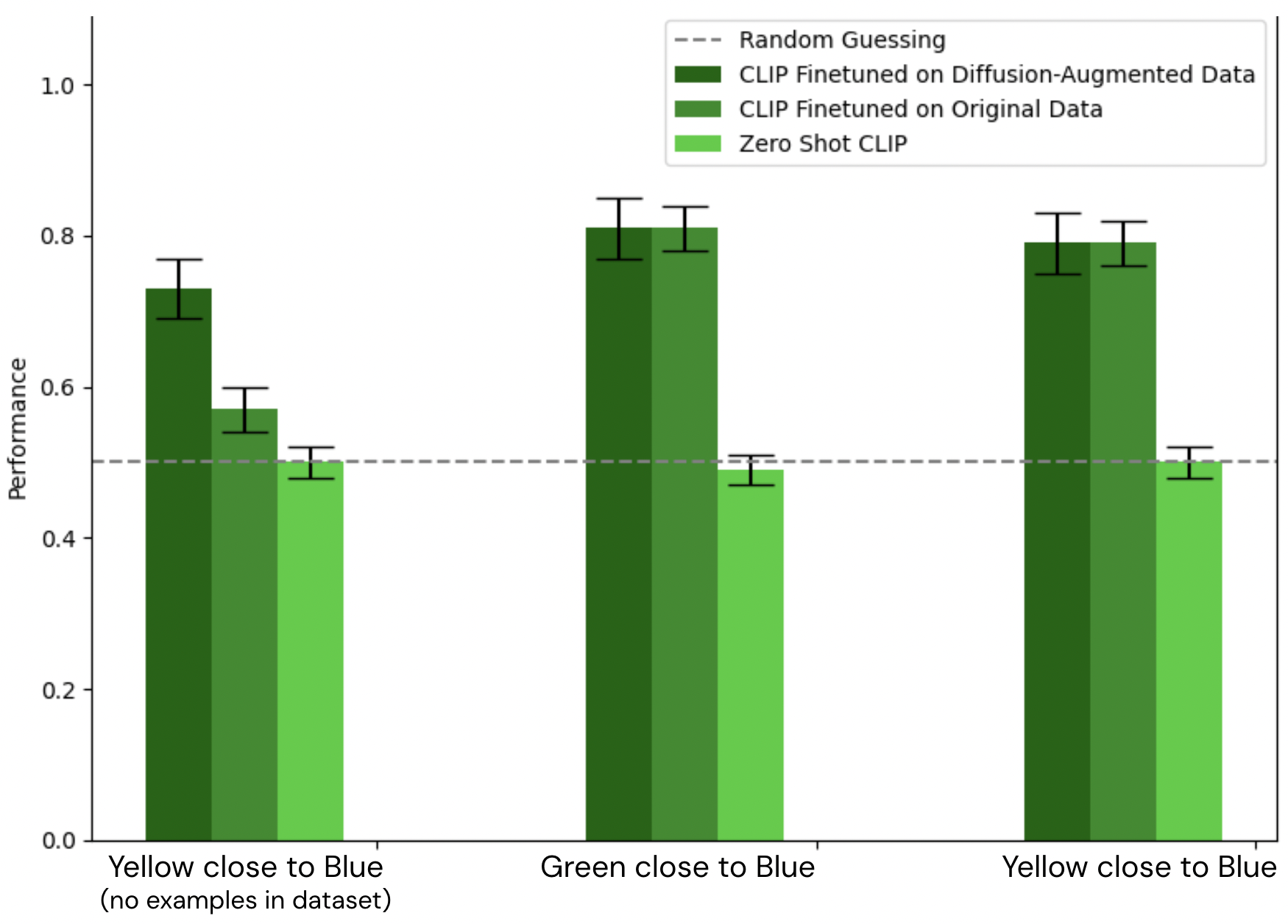}

    \centering
    \caption{Performance of a finetuned CLIP as a reward detector, evaluating the use of synthetic observations to detect reward of a new, unseen task. In all three bar plots, the leftmost task is an held-out task, for which there are no examples in the dataset we use to finetune CLIP. DAAG can synthetically generate data to successfully finetune CLIP as a reward detector, while not affecting the performance on the other tasks. We plot mean and standard deviation by repeating the experiments with three different training and test sets per environment.}
    \label{fig:finetune_vlm}
    \vspace{-15pt}
\end{figure*}

We assume the existence of a dataset $\mathcal{B}$ of collected goal observations for different tasks $\mathcal{T}_\mathcal{B} = [\mathcal{T}_0, \dots, \mathcal{T}_n]$. We then want to measure the performance of CLIP, the VLM we use in this work, to correctly detect novel observations $o_t$ as goal configurations for a new task $\mathcal{T}_i$ not present in $\mathcal{T}_\mathcal{B}$. We compare finetuning CLIP on the original dataset $\mathcal{B}$, and finetuning on an artificially expanded version of $\mathcal{B}$ where we apply diffusion augmentation to synthesise  examples of goal observations for $\mathcal{T}_i$, starting from goal observations of other tasks. 


To empirically evaluate if the synthetic observations positively affect performance, we compare the two pipelines on a test set of unseen observations, where 50\% are goal observations of $\mathcal{T}_i$ and 50\% are not, therefore resulting in a balanced binary classification task. We also compare the zero-shot performance of CLIP, to better evaluate the relative improvement over the off-the-shelf model.

For the \textbf{RGB Stacking} environment, the tasks $\mathcal{T}_\mathcal{B}$ are $[$ \textit{"Stack the green cube on the blue cube"}, \textit{"Stack the blue cube on the red cube"} $]$, respectively $\mathcal{T}_{g,b}^{\text{RGB}}, \mathcal{T}_{b,r}^{\text{RGB}}$, with the test task $\mathcal{T}_i$ being \textit{"Stack the red cube on the green cube"}, $\mathcal{T}_{r,b}^{\text{RGB}}$.

For the \textbf{Room} environment, the tasks are $[$ \textit{"Put a lemon on a red chair"}, \textit{"Put a banana on a blue chair"} $]$, respectively $ \mathcal{T}_{\text{lemon},\text{red}}^{\text{Room}}, \mathcal{T}_{\text{banana},\text{blue}}^{\text{Room}}$, with the test task $\mathcal{T}_i$ being \textit{"Put an apple on a gray chair"}, $\mathcal{T}_{\text{apple},\text{gray}}^{\text{Room}}$.

For the \textbf{Language Table} environment, the tasks are $[$ \textit{"Put the green block near the blue block"}, \textit{"Put the yellow block near the blue block"} $]$, respectively $ \mathcal{T}_{g,b}^{\text{LT}}, \mathcal{T}_{y,b}^{\text{LT}}$, with the test task $\mathcal{T}_i$ being  \textit{"Put the red block near the blue block"}, $\mathcal{T}_{y,b}^{\text{RGB}}$. 

\begin{wrapfigure}{r}{0.3\textwidth}
\vspace{-10pt}
    \begin{center}
    \includegraphics[width=0.3\textwidth]{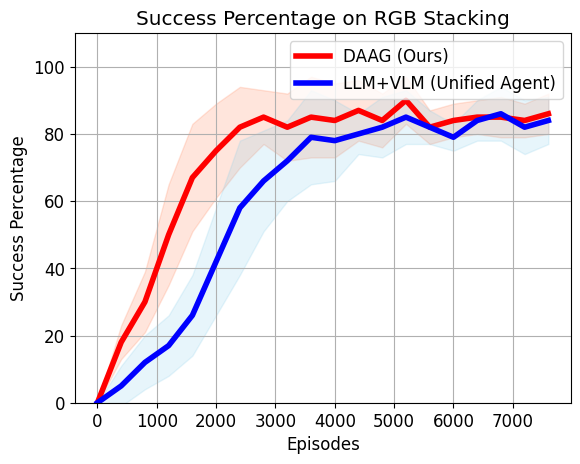}
    \end{center}
    \begin{center}
    \includegraphics[width=0.3\textwidth]{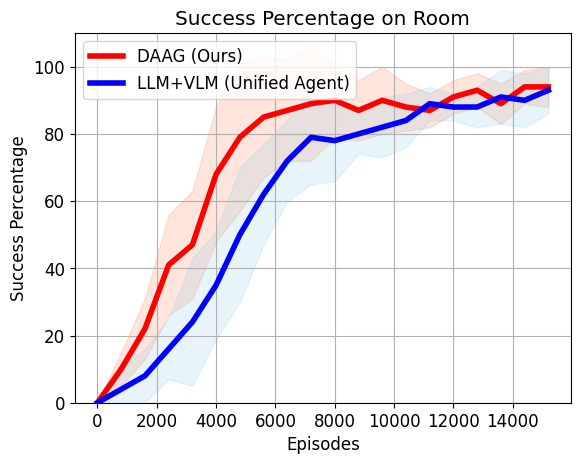}
    \end{center}
    \caption{Performance of learning new tasks from scratch on RGB Stacking and Room. In the plot we show mean and standard deviation over 3 seeds.}
    \label{fig:explore_exp}
    \vspace{-20pt}
\end{wrapfigure}

For each environment, we have $N = 100$ example observations for each $\mathcal{T}_\mathcal{B}$, and use the DAAG pipeline to obtain synthetic observations of $\mathcal{T}_i$ by augmenting all other observations. We finetune CLIP on both the original and augmented datasets and test for accuracy on a test test of $M = 100$ unseen observations. We report results in Figure \ref{fig:finetune_vlm}. The results demonstrate how, in each environment, DAAG can learn an effective reward detector even when having no example observations of such task, \textit{outperforming a CLIP model trained on the other tasks and queried to generalise zero-shot to the new task}. Figure \ref{fig:finetune_vlm} shows how, on the leftmost task that has no examples in the dataset, DAAG brings a substantial improvement by synthesising examples from other tasks, while keeping the same performance on the seen tasks. In the RGB Stacking and Language Table environments, where precise geometric relations between objects poses are fundamental, the difference with the baselines is more impressive, shedding light on the need for diffusion augmentation to obtain an effective reward detector. In the Room environment, the observations CLIP receives, albeit coming from a low-fidelty simulator and renderer, are closer to the distribution of observations it received during training on a web-scale dataset (pictures of fruits and furniture). Therefore, zero-shot performance is considerably stronger there, while in the other tasks is close to random guessing, demonstrating the need for finetuning.

\vspace{-5pt}
\subsection{Can DAAG Explore and Learn New Tasks More Efficiently?}
\vspace{-4pt}

\label{sec:exp_extract}
We here focus on investigating the benefits brought by DAAG and HEA to exploring and learning new tasks from scratch.

We assume the agent start learning a new task \textit{tabula rasa} in this experimental scenario, with an empty new task buffer  $\mathcal{B}_{n}$, and with no access to a lifelong buffer $\mathcal{B}_{ll}$, to independently evaluate the effect of HEA on new task learning efficiency. The agent receives a natural language instruction $\mathcal{T}_i$, that informs it about the goal to achieve in the environment. We experimentally evaluate the learning efficiency improvements brought by HEA, comparing against \citep{dipalo2023unified}, that decomposes $\mathcal{T}_i$ into subgoals $\mathcal{T}_{0:G}$ and obtained reward for each via a VLM, and to a baseline agent that does not benefit neither from task decomposition or diffusion augmentation.

We evaluate the performance of this method on the RGB Stacking and Room environments. For the \textbf{RGB Stacking} environment, the task is  \textit{"Stack the red cube on the blue cube"}, or  $\mathcal{T}_{r,b}^{\text{RGB}}$. For the \textbf{Room} environment, the task is \textit{Put an apple on a gray chair}, $\mathcal{T}_{a,g}^{\text{Room}}$. The environments only provide a reward of $+1$ when the task is successfully achieved, and end the episode there: in that case, we add the entire episode to $\mathcal{B}_{n}$. If an internal reward is detected via the VLM at timestep $T_g$, the observations and actions up to that timestep are added to $\mathcal{B}_{n}$. We perform exploration in the environments via an epsilon-greedy strategy \citep{mnih2013playing}. We start with $\epsilon = 0.99$ and decay it over time, with detailed hyperparameters described in the Supplementary Material. During each episode, if we sample a number $n > \epsilon$ where $n \sim \mathcal{U}(0,1)$, we let the policy network guide the agent. Otherwise, we perform exploration as follows: for the RGB Stacking task, we sample a random pick up position $a_0 \in \mathbb{R}^2$ and a random place position $a_1 \in \mathbb{R}^2$ and execute the actions. For the Room task, in order to speed up and guide exploration, we select a random pickable object in the room, and move the agent to it to pick it up, collecting all observations and actions leading to it $[o_t, a_t]_{t=0}^{T_\text{pick}}$, with $a_t \in \mathbb{R}^2$. We then select a random target furniture where to place the object, and move the agent there collecting other observations and actions $[o_t, a_t]_{t=T_{\text{pick}}}^{T_\text{place}}$. While speeding up the learning of the task, this guidance does not affect the relative performance improvements brought by one method over the other: in the Supplementary Material, we also show training curves with entirely random exploration.   We train our policy every $T_{\text{BC}}$ steps on the observation-action pairs collected in $\mathcal{B}_n$.

In Figure \ref{fig:explore_exp}, we plot the number of successfully solved instances of the task over 100 test episodes as a function of the number of training episodes. During testing, we do not perform any exploration strategy or guidance, and let the policy network guide the agent. We can see how DAAG learns faster than the baseline. The ability to use even certain unsuccessful episodes as learning signal helps improving the learning efficiency across all tested environments.

\subsection{Can DAAG More Effectively Learn Tasks in Succession Transferring Experience from Past Tasks?}
\label{sec:exp_extract}

We now investigate the influence of DAAG on another fundamental ability of lifelong learning agents: the ability to extract, transfer and repurpose past experience to speed up learning of new tasks.  
As we demonstrated the ability of DAAG to learn new tasks efficiently through exploration and HEA in the previous set of experiments, we now investigate its ability to extract information and learn policies from a given dataset of experience, with no additional exploration, in a setting closer to Offline Reinforcement Learning \citep{kostrikov2021offline}. 

\begin{wrapfigure}{r}{0.32\textwidth}
\vspace{-10pt}
    \begin{center}
    \includegraphics[width=0.32\textwidth]{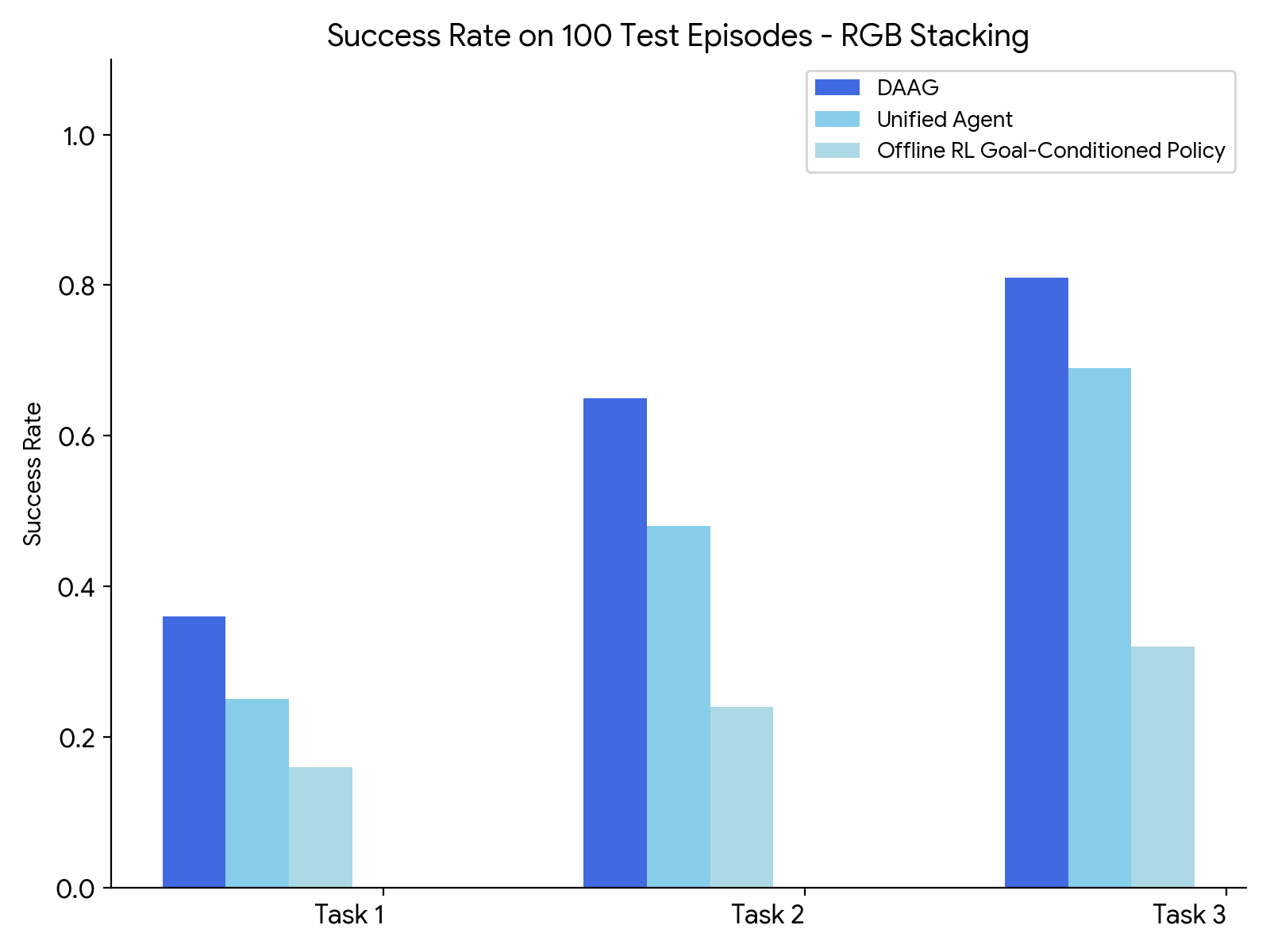}
    \end{center}
    \begin{center}
    \includegraphics[width=0.32\textwidth]{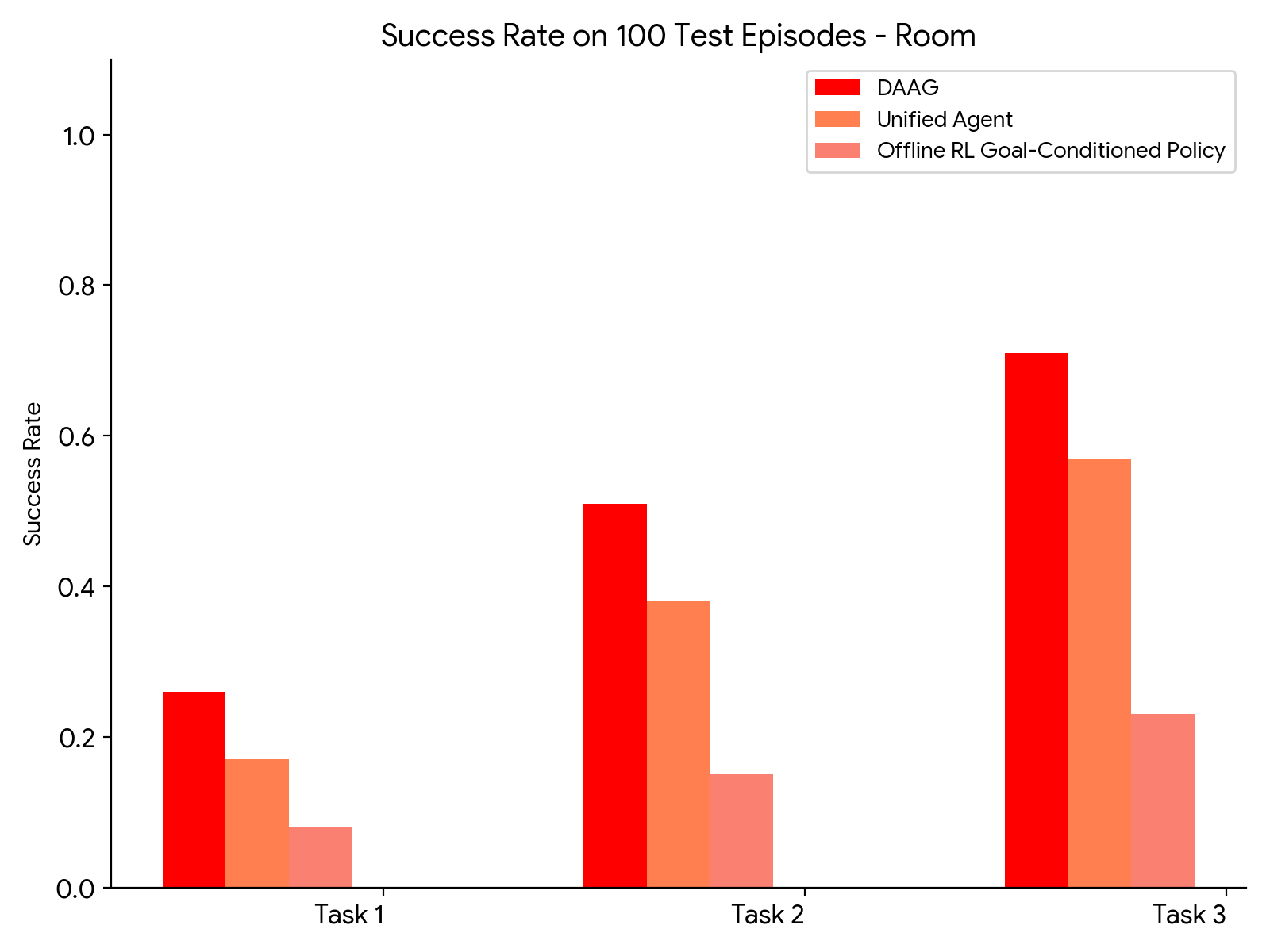}
    \end{center}
    \caption{Sequential task learning performance. By learning to repurpose also episodes solving different but related tasks via HEA, DAAG shows superior transfer learning and lifelong learning performance.}
    \label{fig:extract_exp}
    \vspace{-40pt}
\end{wrapfigure}

We let an agent learn three tasks in sequence per environment. At the beginning of each new task $\mathcal{T}_n$, the agent receives a buffer of experience $\mathcal{B}_{ll,n}$ containing $N_\text{off} = 200$ episodes composed as follows: 50\% are successful episodes of the task at hand $\mathcal{T}_n$, while the other 50\% are episodes solving different tasks in the same environment. Each baseline uses all buffers and data received up to that point $\mathcal{B}_{ll,0:n}$ to learn a policy and is then tested on 100 test episodes to solve the task $\mathcal{T}_n$: for task $n$ the agent has therefore access to $n \times N_{\text{off}}$ pre-collected episodes. 

When learning to solve task $\mathcal{T}_n$ using $\mathcal{B}_{ll,0:n}$, \citep{dipalo2023unified} decomposes $\mathcal{T}_n$ into subgoals $\mathcal{T}_{0:G}$ and for each extracts successful trajectories from $\mathcal{B}_{ll,0:n}$. DAAG, in addition to this, runs \textit{Hindsight Experience Augmentation} to also extract trajectories completing similar tasks that can be visually modified to match any of the subgoals in $\mathcal{T}_{0:G}$. Both baseline train the Self-Imitation Learning policy on the extracted (and synthetically augmented) episodes. In addition, we run another baseline which does not perform any decomposition or extraction, and only trains a goal-conditioned policy on all the episodes contained in $\mathcal{B}_{ll,0:n}$.

For each environment, we learn three tasks in succession. For the RGB Stacking environment, the tasks are, in order, $[$ \textit{"Stack the red cube on the green cube"}, \textit{"Stack the green cube on the blue cube"}, \textit{"Stack the blue cube on the red cube"} $]$, respectively $\mathcal{T}_{r,g}^{\text{RGB}}, \mathcal{T}_{g,b}^{\text{RGB}}, \mathcal{T}_{b,r}^{\text{RGB}}$. 

For the \textbf{Room} environment, the tasks are $[$ \textit{"Put a lemon on a red chair"}, \textit{"Put a banana on a blue chair"}, \textit{"Put an apple on a gray chair"} $]$, respectively $ \mathcal{T}_{\text{lemon},\text{red}}^{\text{Room}}, \mathcal{T}_{\text{banana},\text{blue}}^{\text{Room}}$, $\mathcal{T}_{\text{apple},\text{gray}}^{\text{Room}}$.

In Figure \ref{fig:extract_exp}, we compare the performance, as success rate, of each method on method $\mathcal{T}_n$ using $\mathcal{B}_{ll,0:n}$. We can see how DAAG surpassess both baselines, thanks to the ability to learn from most of the experience stored in $\mathcal{B}_{ll}$, by modifying and repurposing trajectories solving tasks beyond $\mathcal{T}_n$ or its subgoals $\mathcal{T}_{0:G}$.


\begin{figure*}[]
    \centering
    \includegraphics[width=.99\textwidth]{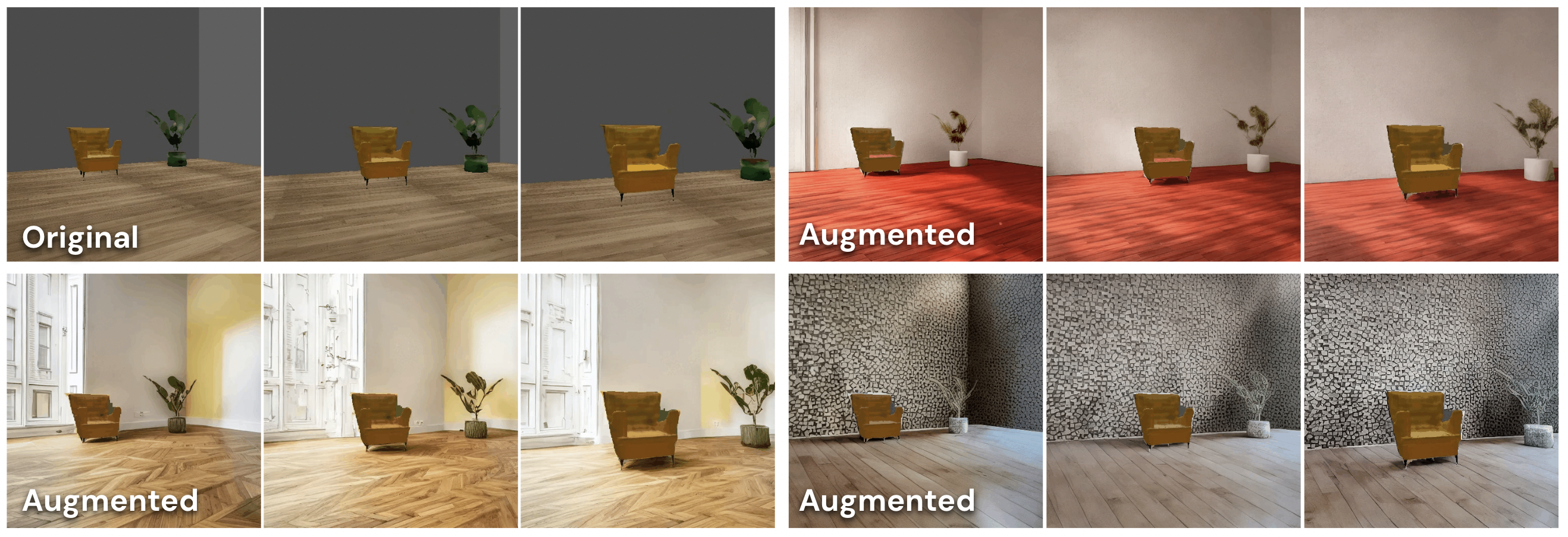}
    \centering
    \caption{Examples of original observations from the Room environment and augmentations obtained through our geometrically and temporally consistent diffusion pipeline.}
    \label{fig:augmented_rooms}
\end{figure*}

\vspace{20pt}

\subsection{Improving Robustness via Scene Visual Augmentation}

\begin{wrapfigure}{r}{0.3\textwidth}
    \begin{center}
    \includegraphics[width=0.3\textwidth]{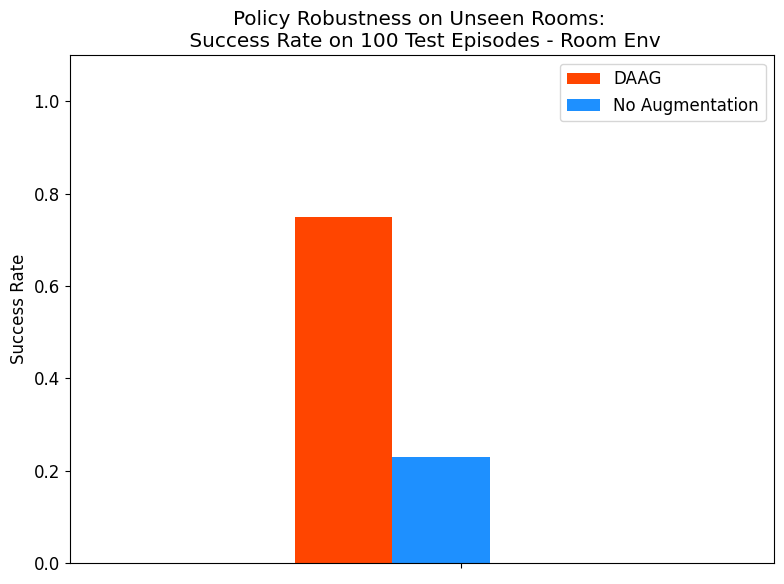}
    \end{center}

    \caption{Success rates of baseline policy and a policy trained on augmented dataset (\ref{fig:augmented_rooms}).}
    \label{fig:aug_rooms_results}
    \vspace{-20pt}
\end{wrapfigure}

We demonstrate how our diffusion pipeline can also be used to visually augment visual observations by modifying the scene and keeping the salient objects untouched, therefore generating additional examples of successful trajectories with different backgrounds. This is possible thanks to both the geometrical consistency and temporal consistency, absent in methods like \cite{chen2023genaug, mandi2023cacti, yu2023scaling}. To test how this affects policy robustness, we gather a dataset of 300 successful episodes in the Room environment where an agent reaches the yellow chair.

We then use our pipeline to augment each observations 5 times, querying the LLM to propose a description of an augmentation (e.g. \textit{a room with a red floor and white walls}). We add all these augmented observations to our buffer and train a policy on it. Both the policy trained on the original and augmented datasets are tested on 5 visually modified room, where we randomly change the walls and floor colors as well as the distractor objects, running 20 test episodes on each room. Figure \ref{fig:aug_rooms_results} demonstrated how visual augmentations leads to a substantially more robust policy, able to reach the target object also on rooms that appear very visually different from the single training room.

\section{Conclusion}
In this work, we proposed Diffusion Augmented Agent (DAAG), a framework that combines large language models, vision-language models, and diffusion models to tackle key challenges in lifelong reinforcement learning for embodied AI agents. 
Specifically, our key results show that DAAG can accurately detect rewards on novel, unseen tasks where traditional approaches fail to generalize. By repurposing experience from prior tasks, DAAG progressively learns each subsequent task more efficiently, requiring fewer episodes thanks to transfer learning. Finally, by diffusing unsuccessful episodes into successful trajectories for related subgoals, DAAG substantially improves exploration efficiency.
Through diffusion augmentation, experience gathered across a lifetime of learning can be repurposed to make each new task easier than the last. This work suggests promising directions for overcoming data scarcity in robot learning and developing more generally capable agents.

\newpage

\section{Acknowledgments}
The authors would like to thank Dushyant Rao for his valuable feedback on earlier drafts of the paper.

\bibliographystyle{abbrvnat}
\bibliography{main}


%

\newpage
\section{Appendix}

\begin{figure*}[t!]
    \centering
    \includegraphics[width=.49\textwidth]{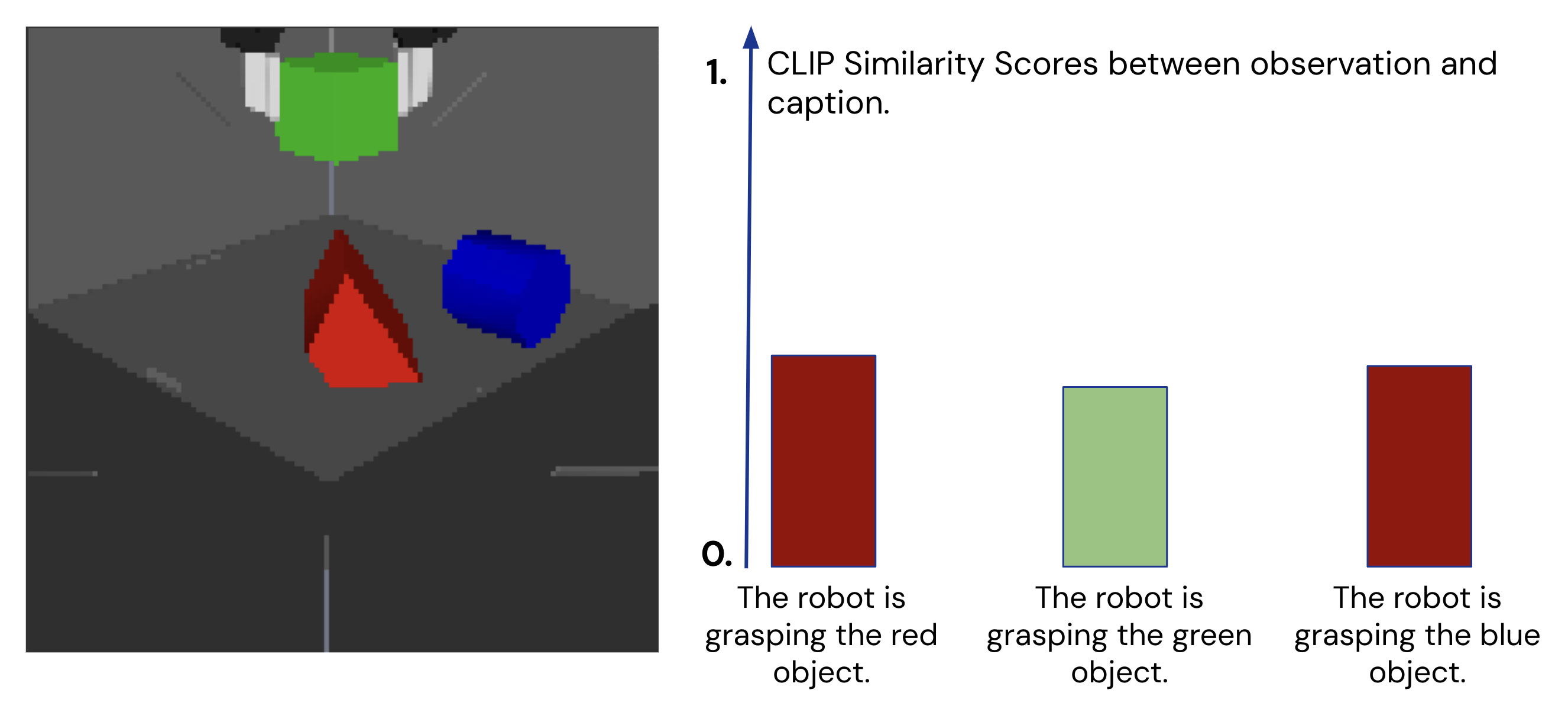}
    \includegraphics[width=.49\textwidth]{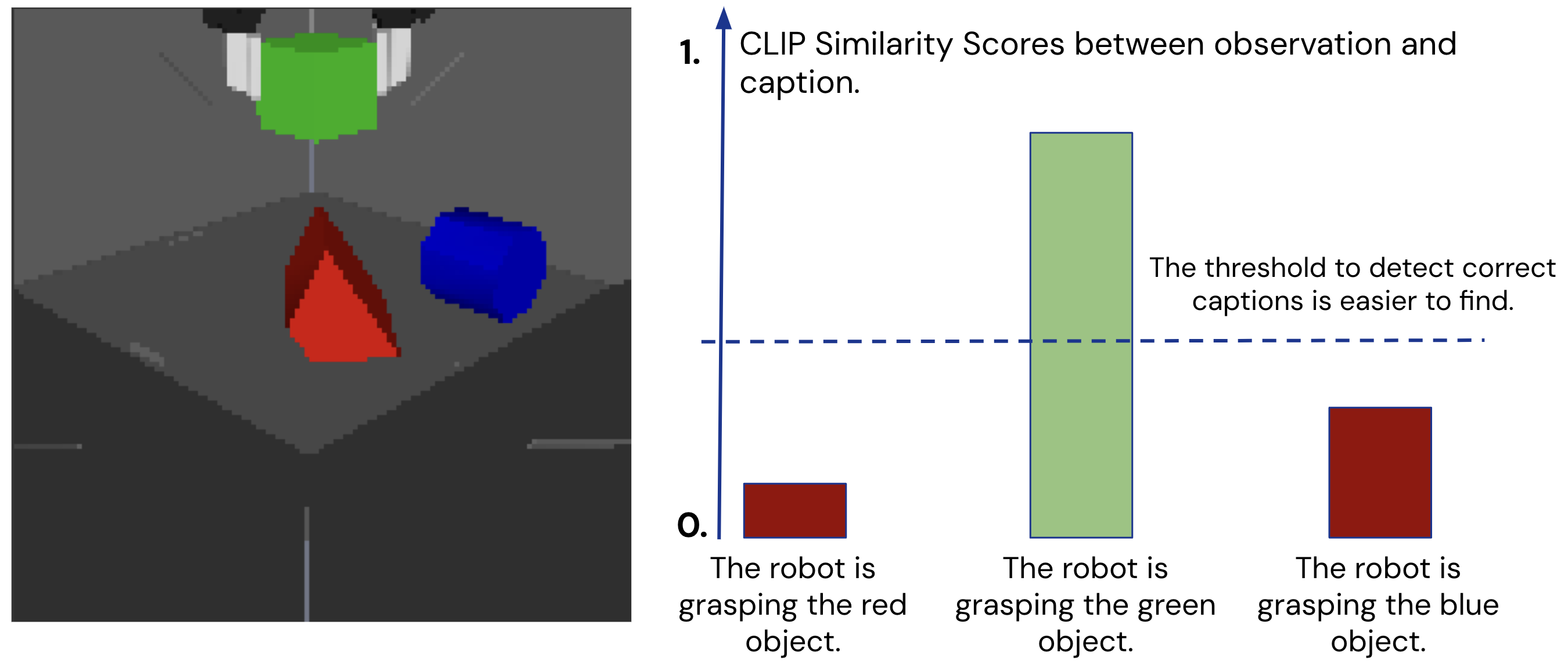}
    \centering
    \caption{Illustrative comparison between the outputs of CLIP before and after finetuning, and how finetuning makes it possible to find a threshold to discriminate correct and wrong text-image matches.}
    \label{fig:clip_before_after}
\end{figure*}

\subsection{Backward Transfer with Hindsight Experience Augmentation}
\begin{wrapfigure}{r}{0.3\textwidth}
    \begin{center}
    \includegraphics[width=0.3\textwidth]{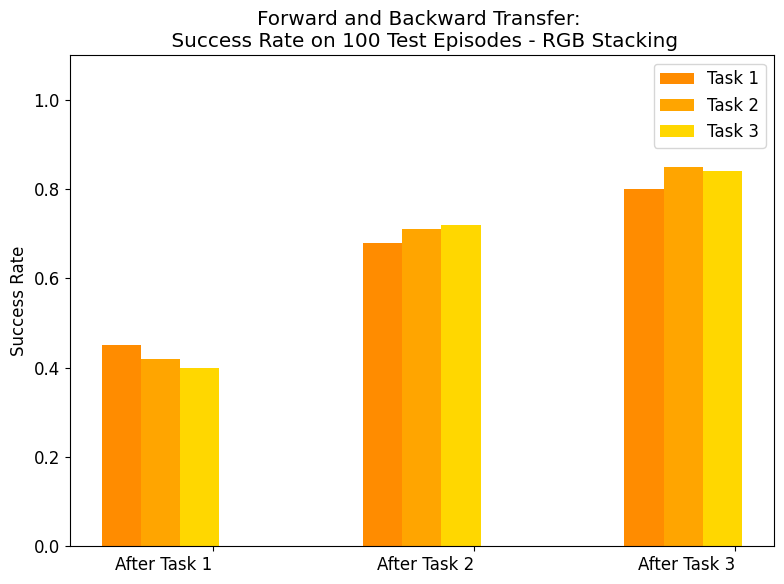}
    \end{center}

    \caption{Backward transfer in a lifelong learning scenario using HEA.}
    \label{fig:forward_backward}
\end{wrapfigure}

In the experiments of section \ref{sec:exp_extract} we studied forward transfer in a lifelong learning setting. We also analysed backward transfer in the RGB Stacking task as follows. After each task of Figure \ref{fig:extract_exp}, we also test performance on the previous tasks by running HEA on all observations gathered up to that points to synthesise additional examples for previous tasks, and retrain and test the policy. Figure \ref{fig:forward_backward} demonstrates how HEA unlocks also strong backward transfer on the same three tasks order as Fig. \ref{fig:extract_exp}: as the agent receives new data for new tasks, HEA can augment this data also to improve performance on previous tasks if needed.

\subsection{Performance with Purely Random Exploration on RGB Stacking}
In our main experiments we used guided exploration to speed up the experiments. However, guided exploration does not change the relative performance of the various methods. We demonstrate how DAAG learns tasks faster than the chosen baseline by re-running the experiment in \ref{fig:explore_exp} (top), using entirely random exploration, where the agent samples a random position to pick and place the objects.

\begin{wrapfigure}{r}{0.3\textwidth}
    \begin{center}
    \includegraphics[width=0.3\textwidth]{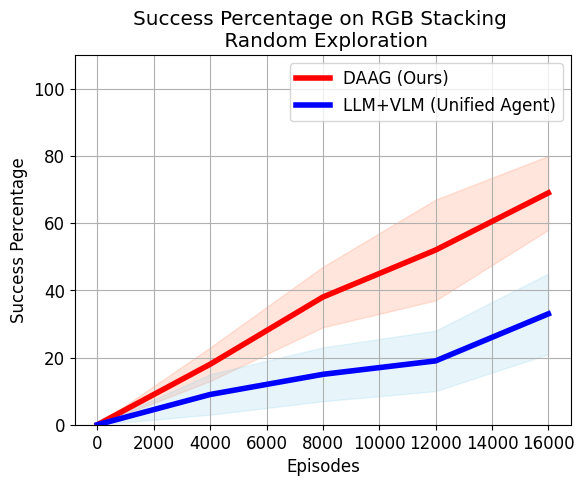}
    \end{center}

    \caption{Performance of DAAG against the baseline with purely random exploration.}
    \label{fig:pure_random}
\end{wrapfigure}

Figure \ref{fig:pure_random} demonstrates how DAAG can learn to tackle a task considerably faster than the baseline that does not use HEA also in this scenario.

\begin{wrapfigure}{r}{0.4\textwidth}
    \begin{center}
    \includegraphics[width=0.4\textwidth]{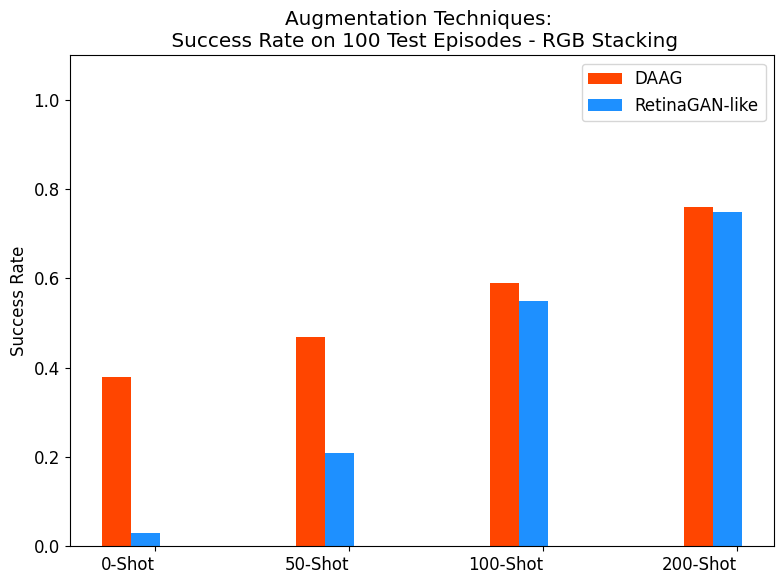}
    \end{center}

    \caption{Performance of DAAG against a RetinaGAN-like style augmentation technique to learn a new task given successful examples of a different tasks and varying amounts of successful examples of the new task at hand.}
    \label{fig:aug_techniques}
\end{wrapfigure}

\subsection{Finetuning CLIP and Finding a Threshold}
As a contrastive VLM, CLIP outputs a similarity score between a textual description and an image. In our work, as also found in \citep{dipalo2023unified}, we observed that off-the-shelf CLIP struggles at recognising precise configurations of objects, performing close to random guessing. 
We therefore finetune it as described in the main paper, increasing the difference in score between correct text-image matches and wrong matches. Given the dataset we used for finetuning, we then find $\delta$, the threshold to detect if a match is correct by comparing if the similarity score is larger, simply by finding the value that would maximise accuracy in an held-out validation set taken from the training set. In Figure \ref{fig:clip_before_after} we illustrate this behaviour.

\subsection{Comparing Different Augmentation Techniques}
In this work, we proposed a diffusion pipeline to visually modify and augment observations, focusing on geometrical and temporal consistency. Here we compare our pipeline with another technique from the recent literature, a CycleGAN-based pipeline \cite{zhu2020cyclegan} inspired by RetinaGAN \cite{ho2021retinagan}. We compare our technique and a RetinaGAN-like technique on the RGB Stacking environment. In particular, we collect 300 successful episodes for the task "\textit{Stack Blue on Green}". We then use both techniques to augment this dataset to a new task, "\textit{Stack Red on Blue}". We compare the performance when receiving 0, 50, 100 or 200 successful examples of the new task. While we can use off-the-shelf Diffusion Models, trained on web-scale datasets, we need to train the CycleGAN on the data from the new task and old task in order to learn a visual mapping. Due to this, DAAG achieves strong 0-shot forward transfer, and keeps improving, while the RetinaGAN-like baseline needs 100 successful examples of the new task to properly learn a visual mapping and re-use data from the old task for the new one, as can be seen in Figure \ref{fig:aug_techniques}.

\subsection{Design of LLM Prompt}
\begin{figure*}[]
    \centering
    \includegraphics[width=.99\textwidth]{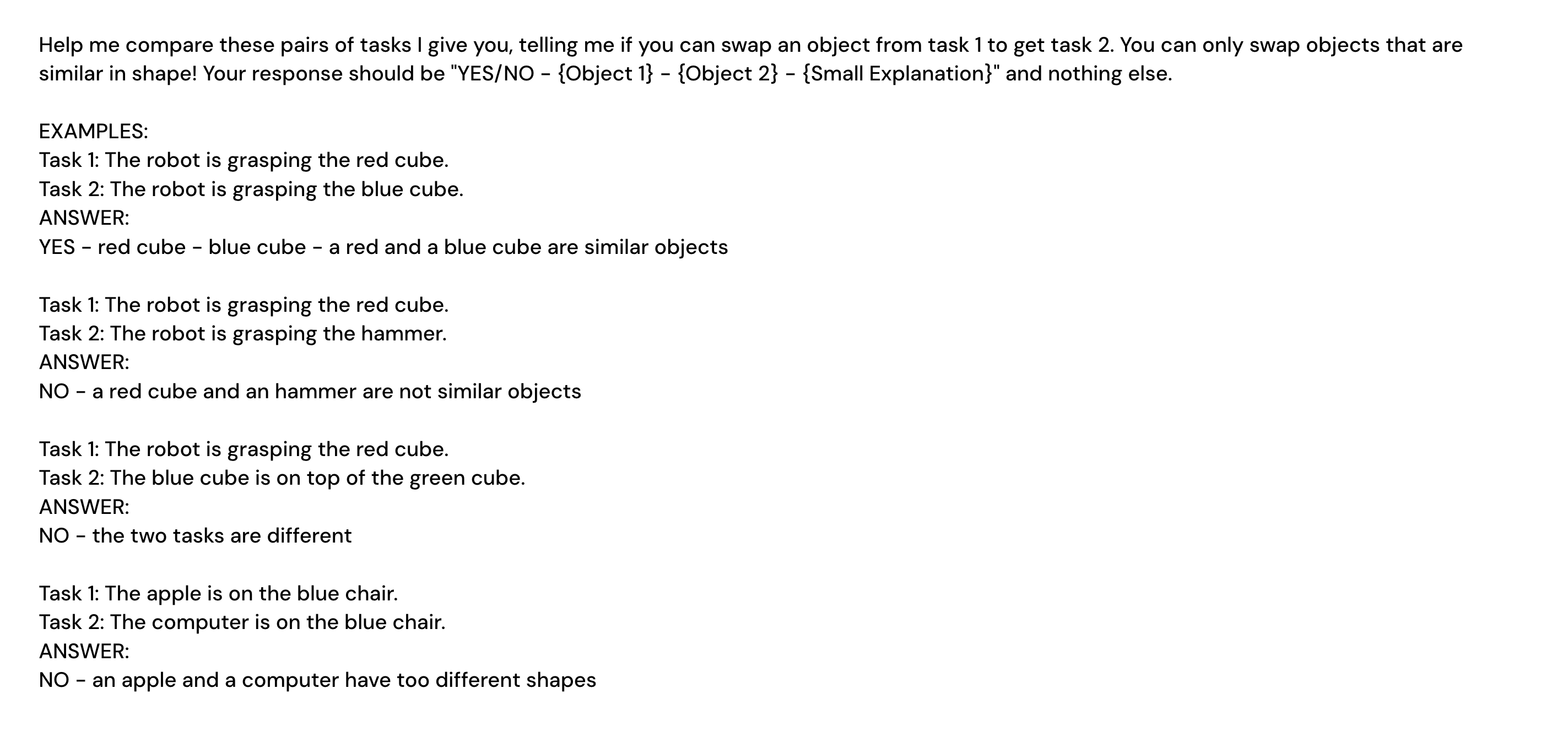}
    \includegraphics[width=.99\textwidth]{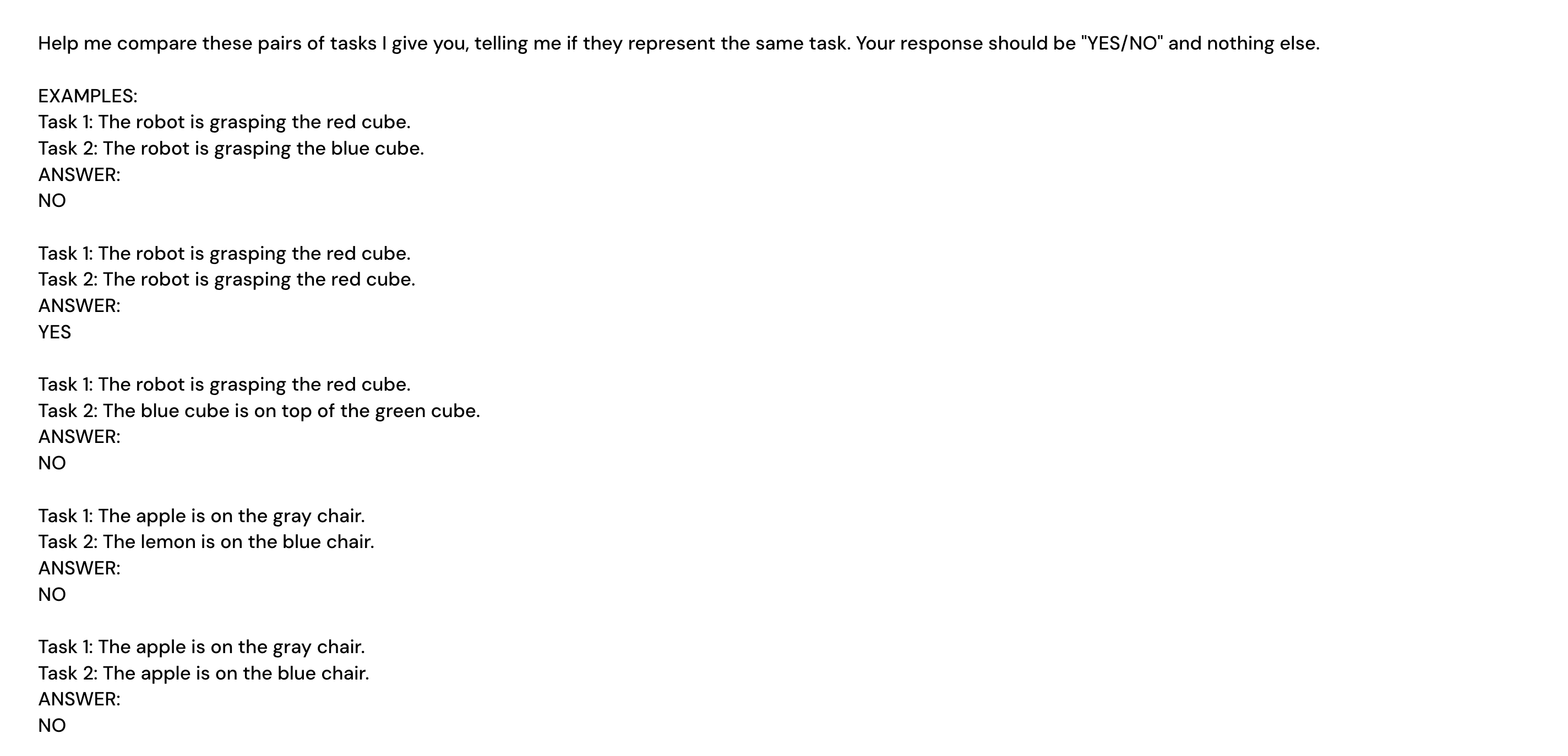}
    \centering
    \caption{Prompts used to detect if a task can be visually modified into another by swapping objects.}
    \label{fig:prompts}
\end{figure*}

In this work, the LLM has two main roles: dividing tasks into subgoals, and comparing tasks to detect if one can be visually modified into the other. For the former, we follow the same prompt proposed in \citep{dipalo2023unified}. For the latter, we illustrate the prompts we used in Figure \ref{fig:prompts}. We use a two stages approach: first, the LLM proposes a possible swap in a format that is easy to parse. Then we apply the swap to the task strings, and compare it again to reduce possible errors.

\subsection{Models Used to Compute the Inputs to ControlNet}
 As visual inputs to ControlNet, we provide, given the original RGB observation, 1) a \textit{canny edges} image that is obtained via OpenCV 2) a \textit{depth image} that we compute using the off-the-shelf model MiDaS \citep{ranftl2020robust} 3) a normals map that is computed algorithmically from the aforementioned depth map.

 All these inputs guide the generation process, and we use a weight of $0.8$ for each in the ControlNet pipeline, using the Diffusers library \href{https://huggingface.co/docs/diffusers/en/index}{https://huggingface.co/docs/diffusers/en/index}.

\newpage

\subsection{Models and Hyperparameters}

\textbf{Large Language Model}: Gemini Pro 1.0

\textbf{Vision Language Model}: CLIP ViT-B/32

\textbf{Diffusion Model}: Stable Diffusion 1.5 with ControlNets (Canny, Depth, Normals)

\textit{Hyperparameters}: Image generation size: $512 \times 512$. Weights: $[0.5,0.8,0.8]$, Generation steps: 20, Guidance Scale = 7

\textbf{Policy Network}: ResNet-18 + 2-layer MLP 

\textbf{Super Resolution Model}: Stable Diffusion 4x Upscaler

\textbf{Object Detection Model}: OWL-ViT

\textbf{Segmentation Model}: FastSAM + CLIPSeg

\textbf{Reinforcement Learning Hyperparameters}: $\epsilon = 0.99$, \text{decay:} 0.9995 \text{ per episode}

\textit{Policy Training}: \textit{batch size} 32, \textit{optimiser} Adam, \textit{learning rate} $1e-4$, \textit{epochs} 5000.

\end{document}